\documentclass[]{bytedance_seed}

\usepackage[toc,page,header]{appendix}

\usepackage{minitoc}
\usepackage{amsfonts}
\usepackage{longtable}
\usepackage{textcomp}
\usepackage{xspace}

\usepackage{algorithm}
\usepackage{algpseudocode}
\usepackage{amsmath}
\usepackage{amssymb}

\usepackage{mfirstuc}
\usepackage{soul}
\usepackage{colortbl}
\newcommand{\faceid}[1]{\texttt{\textless face\_#1\textgreater}}
\newcommand{\facebox}[1]{\texttt{[face\_#1]}}
\newcommand{\wrong}[1]{\textcolor{red}{#1}}
\newcommand{\missing}[1]{\textcolor{blue}{[\textbf{Missing:} #1]}}

\usepackage{enumitem}
\usepackage{wrapfig}

\newcolumntype{P}{>{\raggedright\arraybackslash}p{\dimexpr\textwidth-2\tabcolsep}}

\def\sysname{TaskMem\xspace} 

\title{Task-Focused Memorization for Multimodal Agents}

\author[1,*]{Tao Zou}
\author[1,*]{Yichen He}
\author[1,2]{Tian Qiu}
\author[1, \dagger]{Yuan Lin}
\author[1]{Hang Li}

\affiliation[1]{ByteDance Seed}
\affiliation[2]{Fudan University}

\contribution[*]{Equal contribution}
\contribution[\dagger]{Corresponding authors}

\abstract{

Long-term memory is essential for multimodal agents to build coherent experience, accumulate world knowledge, and achieve continual learning. However, constructing effective memory goes beyond memory module design and basic requirements such as accuracy and fidelity; the key challenge lies in determining what to memorize. Multimodal agents, such as embodied agents, continuously perceive, reason, and act in real or virtual environments, receiving an unbounded stream of multimodal observations. From this combinatorial explosion of information, an agent must selectively retain content that is relevant to its role in the environment and valuable for future tasks. To bridge this gap, we frame memory generation as a learnable memorization policy and introduce TaskMem (\textbf{Task}-focused \textbf{Mem}orization Policy Learning), a reinforcement-learning-based framework that enables the policy to dynamically adjust its focus to the demands of real tasks encountered in the environment. TaskMem adopts a two-phase training paradigm: Phase One learns how to memorize by optimizing memory quality under fundamental fidelity requirements; Phase Two occurs after deployment, where the agent learns what to memorize by tuning an adapter on its base MLLM, using recent environment tasks to define a reward model that guides the memorization policy toward task-relevant content. To evaluate our approach, we reformulate VideoMME, EgoLife, and EgoTempo into streaming benchmarks that simulate a realistic setting in which an agent processes streaming observations and handles tasks arriving online. To isolate memory assessment, the questions must be answered using only the agent's memory, without access to raw video. Built on Qwen3-VL-30B-A3B, TaskMem improves VQA accuracy by 6.3\%, 7.0\%, and 5.3\% on these benchmarks, respectively.

}

\date{\today}
\correspondence{\email{linyuan.0@bytedance.com}}

\checkdata[Project Page]{\url{https://taskmem.github.io/}}

\begin{document}
\maketitle


\section{Introduction}

Long-term memory is a cornerstone of intelligence~\cite{graves2014neural,park2023generative,hendrycks2025definition}. This principle is especially critical for multimodal agents, such as embodied AI agents~\cite{wang2023describe,yang2024embodied,team2025gemini,fang2025robix}, which perceive and act within physical or virtual environments. These agents must continuously perceive and reason over unbounded streams of visual, auditory, and spatial information in dynamic real-world environments. Long-term memory is therefore essential for maintaining coherence across modalities over time~\cite{DBLP:conf/cvpr/SongCWZZWCG0ZLH24,DBLP:conf/cvpr/0004LJJCSSL24}, accumulating world knowledge~\cite{DBLP:journals/corr/abs-2508-09736,wang2024jarvis}, supporting continual learning~\cite{wang2024comprehensive,wang2025karma}, and improving complex, long-horizon decision-making~\cite{wang2024jarvis,li2024optimus,DBLP:journals/corr/abs-2508-09736}. 

\begin{figure}[t]
    \centering
    \includegraphics[width=0.9\linewidth]{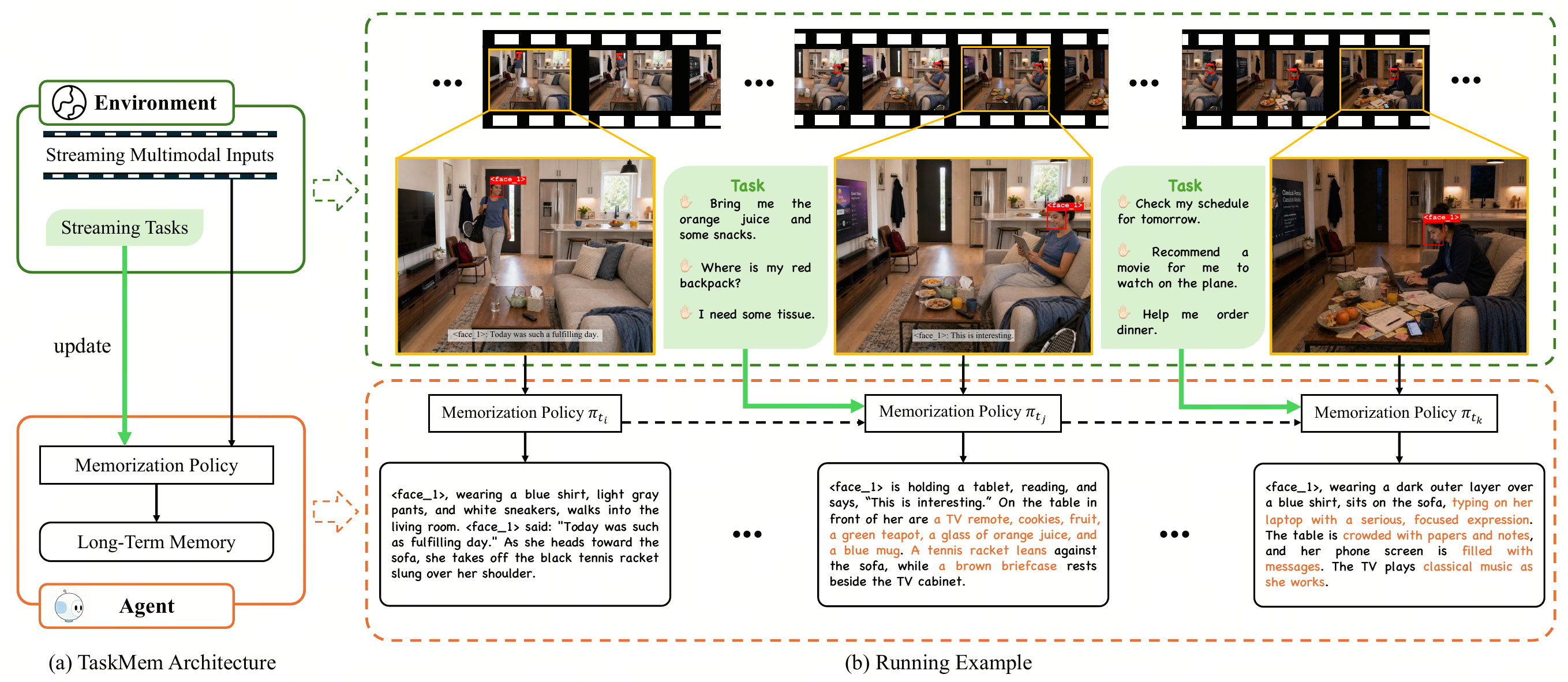}
    \vspace{-5pt}
    \caption{Architecture of \sysname with a running example illustrating its operation. The memorization policy online updates memory generation based on real tasks from the environment, producing task-relevant content.}
    \label{fig:framework}
\end{figure}

The core challenge of memorization lies in whether an agent can \textbf{autonomously decide what to memorize}. Although multimodal agents can perceive and understand vast amounts of information, a fundamental question remains: which information should be stored in long-term memory? This issue also relates to the AI Frame Problem~\cite{shanahan2004frame}, which concerns identifying contextually relevant information without being overwhelmed by the combinatorial explosion of possibilities. Extending this perspective, an agent must not only decide what is relevant in the present, but also what will remain useful in the future. Consequently, memory selection should adapt dynamically to the agent’s role and tasks within its environment. For example, as shown in Figure~\ref{fig:framework}, if a robot is primarily assigned housework tasks, it should focus on constructing memory about the house layout. In contrast, if the robot frequently receives instructions related to its user, it should prioritize building user-centric memories, such as the user's preferences, habits, and emotions. In this sense, memory is not merely a passive storage system, but an active, goal-driven process. An effective agent should continuously retain information that maximizes future utility.

Recently, a growing number of works have proposed various frameworks for multimodal agents with long-term memory~\cite{DBLP:journals/corr/abs-2601-06037,chhikara2025mem0,DBLP:journals/corr/abs-2508-09736,DBLP:journals/corr/abs-2512-02425}. Most of these approaches treat memory construction as an independent process, operating in parallel with task execution. In these frameworks, multimodal large language models (MLLMs)~\cite{comanici2025gemini,xu2025qwen3,bai2025qwen3,singh2025openai,seedseed2,yuan2025tarsier2} are used to generate memory content, alongside system-level mechanisms for memory storage~\cite{DBLP:conf/cvpr/0004LJJCSSL24,DBLP:conf/cvpr/SongCWZZWCG0ZLH24,DBLP:conf/eccv/FanMWDLGL24}, consolidation~\cite{DBLP:journals/corr/abs-2601-06037,DBLP:journals/corr/abs-2512-03627}, and error correction~\cite{DBLP:journals/corr/abs-2602-00415}. 
However, memory content generation itself in existing works remains largely heuristic, relying on prompt engineering~\cite{DBLP:conf/eccv/FanMWDLGL24,DBLP:journals/corr/abs-2601-06037,jiang2026advancing} or post-training with predefined templates~\cite{DBLP:journals/corr/abs-2508-09736,DBLP:journals/corr/abs-2512-03627}, and does not explicitly optimize what information should be memorized. Consequently, the formation of memory itself may not be well aligned with the demands of tasks in the real-world environment. 

To bridge this gap, we frame memory generation as a learnable memorization policy rather than a fixed summarization step. Given streaming multimodal inputs and recent memory history, the policy decides what information to store at each moment. 

As shown in Figure~\ref{fig:framework}, we introduce \sysname, a reinforcement learning (RL)-based framework that optimizes the memorization policy so that generated memories are both intrinsically correct and relevant to tasks in the agent's deployment environment. \sysname adopts a two-phase optimization. In \textbf{Phase One}, the policy learns how to memorize by optimizing memory quality under fundamental requirements such as correctness, non-redundancy, and format compliance, which we formulate as a multi-objective RL problem. In \textbf{Phase Two}, the policy learns what to memorize through online learning in an actual environment. This online setting raises several challenges: (1) sparse feedback, as updates rely on only a small number of recent tasks; (2) catastrophic forgetting of capabilities acquired in Phase One; (3) computational constraints, as updates must not affect serving. To address them, we tune a lightweight adapter with only 2,048 parameters on top of the base MLLM, with a reward model that transforms sparse, task-level signals into denser supervision by constructing augmented pairwise preference data, thereby guiding the policy toward task-relevant memory while preserving the general capabilities learned in Phase One.


We evaluate \sysname by recasting Video Question Answering (VQA) benchmarks, VideoMME~\cite{fu2025video}, EgoLife~\cite{yang2025egolife}, and EgoTempo~\cite{plizzari2025omnia}, into sequential task streams that simulate a multimodal agent perceiving and processing tasks sequentially within an environment. For each benchmark, we group video-question pairs by question type and treat each group as a distinct task, representing a specific environment in which we evaluate whether the agent can generate task-relevant memory. Within a task, videos are presented to the agent in sequence, and the agent generates episodic memory for each video. The corresponding question is revealed only after the video has been processed. To isolate the memory assessment, each question must be answered using only the memory, without access to the original video. The resulting accuracy therefore reflects the quality of the generated memory.

We implement \sysname based on Qwen3-VL-30B-A3B~\cite{bai2025qwen3}. Experiments show that Phase One memory learning alone improves VQA accuracy over the base model. Phase Two training further aligns memory with the environment, yielding consistent gains over the base model and overall improving accuracy by 6.3\%, 7.0\%, and 5.3\% on the three benchmarks, respectively. These results show that task-focused memorization effectively improves memory utility.

The main contributions of this paper are summarized as follows:
\begin{itemize}[itemsep=3pt, topsep=0pt, parsep=0pt]
    \item We frame memory generation as a learnable policy that autonomously decides what to memorize from streaming multimodal inputs, addressing the core challenge of memory selection and transforming memory from passive storage into an active, goal-driven process.

    \item We propose \sysname, an RL–based framework that optimizes the memorization policy to generate task-relevant memory within an environment. 
    \item In streaming VQA experiments, both Phase One and Phase Two training demonstrate consistent improvements across VideoMME, EgoLife, and EgoTempo benchmarks.

\end{itemize}




\section{Approach}\label{sec:approach}


\subsection{Problem Formulation}\label{subsection:problem_formulation}
Following mainstream work on long-term memory~\cite{DBLP:journals/corr/abs-2508-09736,DBLP:journals/corr/abs-2512-02425,yang2025egolife,lin2025hippomm}, we formulate the memorization process as a task that takes a streaming video as input and generates memory content.
As a representative case, we focus on episodic memory, which captures temporally ordered, event-centric experiences of a multimodal agent. The same formulation can extend to semantic and visual memory.

At each step $t$, the agent observes a new video segment $v_t$ and maintains the memories generated so far. The memorization policy conditions on a sliding-window context consisting of the $k$ most recent video segments and the memories generated for the first $k-1$ segments in this window, denoted as $q_t=(v_{t-k+1:t}, m_{t-k+1:t-1})$. Given the context $q_t$, the memorization policy $\pi_\theta(m_t|q_t)$ determines a memory $m_t$ for the segment $v_t$. A good episodic memory is faithful to the current video segment, coherent with previous memories, non-redundant, and useful for future tasks in the environment.

Maintaining consistent identities across video segments requires linking the faces and voices in each clip to individuals seen earlier~\cite{he2024storyteller}. \sysname achieves this by annotating the video input itself with persistent identities: detected faces are enclosed in bounding boxes labeled with global face IDs, and ASR transcripts are overlaid as time-aligned subtitles tagged with speaker IDs. Unlike prior work that appends lengthy tool outputs as additional textual context~\cite{DBLP:journals/corr/abs-2508-09736}, this design shortens the context without sacrificing identity reasoning accuracy. Implementation details are in Appendix~\ref{appendix:tools}.


In RL terminology, the full trajectory of the memorization policy is $\tau=(v_{t-k+1:t}, m_{t-k+1:t})$. A trajectory-level reward $r(\tau)$ is provided at the last token of the trajectory. The objective is to learn a memorization policy $\pi_\theta$ that maximizes the expected return: $\mathbb{E}_{\tau\sim\pi_\theta}[r(\tau)]$. For notational simplicity, in the remainder of the paper, we drop the absolute time index and re-index the sliding-window context as
$q=(v_1,\ldots,v_k,m_1,\ldots,m_{k-1})$ and the corresponding full trajectory as $\tau=(q, m_k)$.



\begin{figure}[t]
    \centering
    \includegraphics[width=0.8\linewidth]{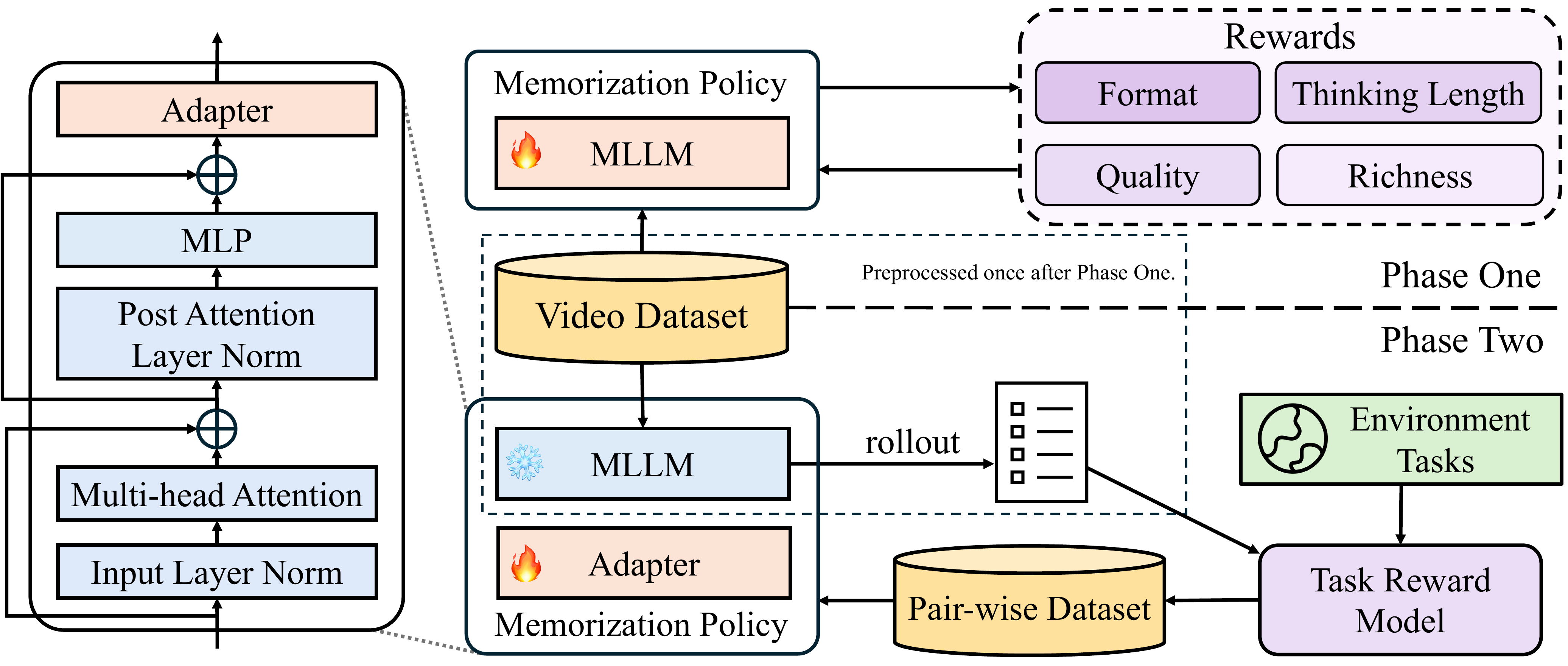}
    
    \caption{Two-phase training in \sysname: Phase One optimizes the memorization policy for fundamental capabilities; Phase Two optimizes the policy to generate task-relevant content.}
    \label{fig:train}
\end{figure}

\subsection{Phase One: How to Memorize}
\label{subsection:phase_one}
Phase One optimization occurs before the agent is deployed in real-world environments. At this stage, we aim to optimize the memorization policy by satisfying fundamental requirements, such as factual accuracy, non-redundancy, and proper formatting. Memory generation is an open-ended, long-form task. Prior work mainly relies on supervised fine-tuning (SFT)~\cite{DBLP:journals/corr/abs-2508-09736,yang2025egolife}, which has two limitations. First, performance is capped by the quality of the models used to generate synthetic training data. Second, the maximum likelihood objective does not explicitly enforce global-level goals. To overcome these limitations, we adopt RL to directly train memory generation from scratch, as shown in Figure~\ref{fig:train}. RL allows direct optimization via reward signals, removing the need for curated SFT data and requiring only raw video data for training. Additionally, large multimodal and language models often have stronger evaluation abilities than generation abilities~\cite{saunders2022self,bai2022constitutional}. For example, the tasks that judge factual alignment with a video or detect redundancy are often easier than producing accurate and non-redundant content. This suggests that an RL-based approach, which leverages these critic strengths as reward signals, can produce a higher-quality memorization policy and provide a strong foundation for future Phase Two online training in the environment.

\textbf{Optimization Algorithm.} We adopt the Group Sequence Policy Optimization (GSPO) algorithm for RL training because of its improved training stability and efficiency in sequence-level reward settings~\cite{zheng2025group,zheng2025stabilizing}. For each training input $q=(v_1, \cdots, v_k, m_1, \cdots, m_{k-1})$, the memorization policy $\pi_\theta$ rolls out a group of $G$ trajectories $\tau_{i=1}^G$, where $\tau_i = (q, m_{k, i})$. The reward for the memory construction of the $i$-th rollout, denoted $r_\text{mc}(\tau_i)$, is given by reward models. Then, we compute the advantage of each trajectory by normalizing rewards within the group:
\begin{equation}
    \hat{A}_i = \frac{r_\text{mc}(\tau_i) - \text{mean}(\{r_\text{mc}(\tau_i)\}_{i=1}^{G})}{\text{std}(\{r_\text{mc}(\tau_i)\}_{i=1}^G)}.
\end{equation}
The optimization objective is:
\begin{equation}
    \mathcal{J}_\text{GSPO}(\theta)=\mathbb{E}_{q\sim\mathcal{D}, \{\tau_i\}_{i=1}^G\sim\pi_{\theta_\text{old}}(\cdot|q)}\left[\frac{1}{G}\sum_{i=1}^{G}\text{min}\left(s_i(\theta)\hat{A}_i, \text{clip}(s_i(\theta), 1-\epsilon, 1+\epsilon)\hat{A}_i\right)\right],
\end{equation}
where the importance ratio $s_i(\theta)$ is defined as 
$s_i(\theta) = \left(\frac{\pi_\theta(m_{k,i}|q)}{\pi_{\theta_\text{old}}(m_{k,i}|q)}\right)^{\frac{1}{|m_{k,i}|}}.$

\textbf{Multi-Objective Reward Design.}
We define the trajectory-level reward $r_{\mathrm{mc}}(\tau)$ as the sum of four components:

\begin{equation}
r_{\mathrm{mc}}(\tau)
= r_{\mathrm{fmt}}(\tau) + r_{\mathrm{len}}(\tau) + r_{\mathrm{qual}}(\tau)
+
r_{\mathrm{rich}}(\tau).
\end{equation}

Here, $r_{\mathrm{fmt}}$ evaluates whether the output follows the predefined format. Following the ReAct~\cite{yao2023react} paradigm, the policy generates reasoning before memory, and $r_{\mathrm{len}}$ is a soft penalty on overlong intermediate reasoning to regularize computational overhead. $r_{\mathrm{qual}}$ measures the quality of the generated memory in terms of accuracy, non-redundancy, and style, as evaluated by reward models.

Optimizing solely for quality can lead the memorization policy to hack the objective, generating outputs that are accurate but lack substantive content (see examples in Table~\ref{table:case_study_memory_cases}). To address this issue, we introduce a richness reward $r_{\mathrm{rich}}(\tau)$, which explicitly encourages the generation of content-rich memories. Richness is defined relatively within each group $\{\tau_i\}_{i=1}^{G}$. A reward model ranks each sampled memory by richness, and these rankings are then converted into scalar rewards. 

Detailed definitions of all reward components, including prompts and scoring rules, are in Appendix~\ref{appendix:reward_details}.

\subsection{Phase Two: What to Memorize}

When deployed in real-world environments, the agent should generate task-relevant memory. To achieve this, Phase Two employs online learning driven by environment feedback, allowing the agent to periodically update its parameters for timely adaptation. The core intuition mirrors how humans refine behavior based on past experience: the agent leverages recent tasks to model the likely distribution of future tasks and, accordingly, adjusts its memory focus, determining what to memorize in subsequent steps. 
Specifically, within each update window, \sysname utilizes the most recent $n$ tasks to optimize memory generation. 

Phase Two training presents several challenges. (1) Sparse feedback: \sysname must adapt its memorization policy using only a small number of recent tasks (e.g., ten questions), making the learning signal limited. (2) Catastrophic forgetting: Updating the policy parameters risks overwriting or degrading previously learned capabilities. (3) Computational efficiency: Since adaptation occurs at deployment time, updates must be both fast and resource-efficient. In the following, we describe our design and explain how it addresses these challenges.   

\textbf{Feedback Augmentation.} To enrich feedback signals, we introduce a reward model that takes a set of example tasks and two candidate memories as input. The model first infers the underlying intent of the example tasks, and then determines which memory is more relevant to the corresponding environment, or whether both are similarly relevant.

We pre-construct a dataset of rollouts sampled from the Phase One memorization policy. Specifically, for each input $q=(v_{1:k}, m_{1:k-1})$, we roll out a set of candidate memories. At deployment time, this pre-computed dataset allows us to efficiently construct pairwise preference data $(q, m_k^w, m_k^l)$ by leveraging the reward model to compare candidate memories. These pairwise comparisons are then used to optimize the memorization policy, encouraging it to generate more relevant memories.

\textbf{Model Architecture.} To address the issues of catastrophic forgetting and computational efficiency, we adopt a simple parameter-efficient tuning~\cite{hu2023llm,ding2023parameter} approach. Specifically, as shown in Figure~\ref{fig:train}, we introduce an additive trainable vector adapter at a selected transformer layer, defined as
\begin{equation}
h_o \leftarrow h_o + a.
\end{equation}
Here, $h_o\in\mathbb{R}^d$ is the layer output and $a\in\mathbb{R}^d$ is a learnable vector, only inserted at a single layer. 

\textbf{Optimization Algorithm.} After augmenting the task signals to pairwise data, we adopt the Direct Preference Optimization (DPO) algorithm~\cite{rafailov2023direct} for Phase Two training. The training objective is 
\begin{equation}
    \mathcal{L}_\text{DPO}(\pi_\theta;\pi_\text{ref})=-\mathbb{E}_{(q, m_k^w, m_k^l)\sim\mathcal{D}}\left[\log\sigma\left(\beta\log\frac{\pi_\theta(m_k^w|q)}{\pi_\text{ref}(m_k^w|q)}-\beta\log\frac{\pi_\theta(m_k^l|q)}{\pi_\text{ref}(m_k^l|q)}\right)\right].
\end{equation}
Here, $\pi_\text{ref}$ is the policy obtained from Phase One training. During Phase Two, we update only the adapter parameters, while keeping the backbone model fixed.





\section{Experiments}


In this section, we reformulate VQA benchmarks into a streaming setting by grouping QA pairs of the same type into task-specific environments, enabling us to assess whether \sysname can learn to generate task-relevant memory. In the following, we first describe the two-phase training details and present the training dynamics, showing how Phase One establishes basic capabilities and Phase Two learns to produce task-relevant memories. We then introduce the evaluation setup and report VQA performance when questions are answered solely from the generated memory, thereby further demonstrating its effectiveness for solving tasks in the environment.

\subsection{Training}


\subsubsection{Phase One Training}

We fine-tune Qwen3-VL-30B-A3B~\cite{bai2025qwen3} as the memorization policy with GSPO in two stages, which differ in how the historical memories $m_{1:k-1}$ are generated. Since obtaining histories from the current policy is expensive, the first stage uses Gemini-2.5-Pro~\cite{comanici2025gemini} to synthesize trajectories $m_{1:k-1}$ for each video clip $v_{1:k}$ sampled from an in-house long video dataset $\mathcal{D}$ (see prompt in Table~\ref{tab:prompt_generating_episodic_memory}). We refer to this stage as off-policy history training. The synthesized memories are used as history context, while GSPO optimizes the memory $m_k$ generated by the policy. 

This stage leaves a distribution gap: at deployment, all historical memories are generated by the memorization policy itself, whereas during this stage, they come from a different model. To close this gap, we proceed to perform on-policy history training, continuing GSPO with histories generated by the current policy. This aligns the distribution between training-time and test-time.

\captionsetup[algorithm]{font={scriptsize}}
\begin{algorithm}[h]
\scriptsize
\caption{Phase One On-Policy History Training Algorithm}\label{phase_one_training}
\begin{algorithmic}[]
\Require Video dataset $\mathcal{D}$, batch size $B$, keeping probability $p$, maximum video clips $K=5$, clip thresholds $n_\text{min}, n_\text{max}$, training steps $T$.
\Ensure Memory policy $\pi_\theta$.
\State 
\For{$j=1$ \textbf{to} $B$}\Comment{Initialization}
    \State Sample a video $v$ from $\mathcal{D}$ and a random starting point $i$.
    \State $x_j\leftarrow(v_i)$, \quad $c_j\leftarrow1$\Comment{$c_j$: processed clip counter}
\EndFor
\State
\For{$t=1$ \textbf{to} $T$}\Comment{Training loop}
    \For {each $x_j=(v_i, \cdots, v_{i+k}, m_i,\cdots,m_{i+k-1})$ in batch}\Comment{GSPO optimization on current batch}
        \State Sample a group of memory $M_{i+k}^{(j)}=\{m_{i+k,1}, \cdots, m_{i+k,G}\}$.
        \State Compute reward and advantage of each memory in $M_{i+k}^{(j)}$.
    \EndFor
    \State Update policy parameters $\theta$ by GSPO algorithm.
    \State
    \For {each $(x_j, c_j)$ in batch}\Comment{Batch data evolution}
        \If {$c_j<n_\text{min}$}
            \State $\textsc{Extend}(j)$\Comment{Always retain below lower threshold}
        \ElsIf {$c_j>n_\text{max}$}
            \State $\textsc{Resample}(j)$\Comment{Always discard above upper threshold}
        \Else
            \State With probability $p$: $\textsc{Extend}(j)$; otherwise $\textsc{Resample}(j)$
        \EndIf
    \EndFor
\EndFor
\State
\State \textbf{return} $\pi_\theta$

\State

\Function{Extend}{$j$} \Comment{Retain and extend}
    \State Select a $\hat{m}_{i+k}$ uniformly from $M_{i+k}^{(j)}$.
    \If {$k + 1 < K$}
        State $x_j \leftarrow (v_i, \cdots, v_{i+k+1}, m_i, \cdots, m_{i+k-1}, \hat{m}_{i+k})$
    \Else \Comment{Slide window to keep context $\leq K$}
        \State $x_j \leftarrow (v_{i+1}, \cdots, v_{i+k+1}, m_{i+1}, \cdots, m_{i+k-1}, \hat{m}_{i+k})$
    \EndIf
    $c_j\leftarrow c_j + 1$
\EndFunction

\State

\Function{Resample}{$j$}\Comment{Discard and re-sample}
    \State Sample new video $v\sim\mathcal{D}$ and a random starting point $i$.
    \State $x_j\leftarrow(v_i), \quad c_j\leftarrow 1$
\EndFunction
    
\end{algorithmic}
\end{algorithm}
\captionsetup[algorithm]{font={normalsize}}

We summarize the on-policy history training in Algorithm~\ref{phase_one_training}. At initialization, a batch is constructed by sampling videos from $\mathcal{D}$ and randomly selecting one 10-second clip per video as the initial input. At each training step, we sample a group of candidate memory actions for every batch element, score them with reward models, and update the policy via the GSPO loss. After optimization, the batch is adjusted to maintain video diversity and well-distributed training contexts. Each instance tracks a clip counter $c_j$ indicating the number of clips from the same long video it has consumed. Instances with $c_j < n_\text{min}$ are always extended by appending a sampled memory action and the next video clip (with a sliding window ensuring the context contains at most $K=5$  clips). This guarantees sufficient exposure to longer contexts. Instances with $c_j > n_\text{max}$ are discarded and replaced by a newly sampled clip from another long video in $\mathcal{D}$. Instances between the two thresholds are retained with probability $p$, and resampled otherwise.


For the reward model implementation, the format reward $r_\text{fmt}$ and length penalty $r_\text{len}$ are rule-based. The quality reward $r_\text{qual}$ is obtained by prompting Gemini-2.5-Flash and GPT-4o. The richness reward $r_\text{rich}$ is derived by prompting GPT-4o. Additional implementation details are provided in Appendix~\ref{reward_implementation}.

In total, we use 326 long videos, with an average of 25.15 clips per video for training. Table~\ref{tab:gspo_hyperparameters} lists the hyperparameters used during the GSPO training process.




\begin{figure}[th]
    \centering
    \includegraphics[width=1.0\linewidth]{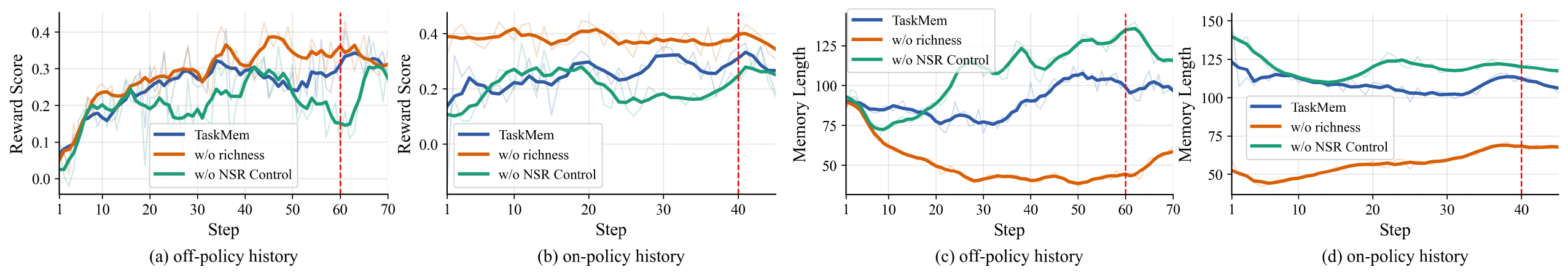}
    \vspace{-5pt}
    \caption{Comparison of training performance under three settings: \sysname, \sysname without the richness reward (w/o richness), and \sysname without NSR control (w/o NSR control). (a) and (b) show reward trajectories during the GSPO training for off-policy history training and on-policy history training, respectively. (c) and (d) show the corresponding changes in memory length over training for off-policy history training and on-policy history training, respectively.}
    \label{fig:training_curve}
\end{figure}

\textbf{Richness Reward Discussion} We compare the training dynamics with and without the richness reward in Figure~\ref{fig:training_curve}. Training without the richness reward appears more stable and achieves a higher reward. However, this does not indicate a better policy; instead, the policy hacks the rewards. As shown in Figure~\ref{fig:training_curve}, the memory length decreases rapidly, suggesting that the policy learns to generate shorter memory sequences to obtain higher quality scores. Table~\ref{table:case_study_memory_cases} further illustrates this behavior with a case where the policy, trained without the richness reward, produces accurate but less substantive content.

\textbf{Stabilizing Training} We extend the concepts of~\cite{zhusurprising} to a multi-valued reward setting by decomposing the learning signal into two components: Positive Sample Reinforce (PSR), which reinforces responses with positive advantage, and Negative Sample Reinforce (NSR), which penalized those with negative advantage. 


We observe that once the policy's average reward reaches a relatively high level, NSR begins to introduce instability, as shown in Figure~\ref{fig:training_curve}. To address this, we adopt a simple yet effective strategy: for samples with positive reward, NSR is disabled, improving training stability. Specifically, the training objective becomes:

\begin{equation}
\scriptsize
    \mathcal{J}_\text{GSPO}(\theta)=\mathbb{E}_{q\sim\mathcal{D}}\frac{1}{G}\left[\underbrace{\sum_{\hat{A}_i>0}\text{min}\left(s_i(\theta)\hat{A}_i, \text{clip}(s_i(\theta), 1-\epsilon, 1+\epsilon)\hat{A}_i\right)}_{\text{Positive Sample Reinforce}} + \underbrace{\sum_{\hat{A}_i<0}\mathbb{I}(r_\text{mc}(\tau_i))\cdot\text{min}\left(s_i(\theta)\hat{A}_i, \text{clip}(s_i(\theta), 1-\epsilon, 1+\epsilon)\hat{A}_i\right)}_\text{Negative Sample Reinforce}\right],
\end{equation}
where $\mathbb{I}(r_\text{mc}(\tau_i))=1$ if $r_\text{mc}(\tau_i) < 0.0$, and $\mathbb{I}(r_\text{mc}(\tau_i))=0$ otherwise.



\subsubsection{Phase Two Training}

We first construct a dataset of (video, rollouts) pairs, where rollouts are sampled using the memorization policy $\pi_0$ obtained from Phase One training. During deployment, the agent collects real tasks and leverages the reward model to transform sparse feedback into pairwise preference data. We then perform DPO on this dataset, training only a lightweight adapter while keeping the base MLLM frozen. Training hyperparameters are listed in Table~\ref{tab:dpo_hyperparameters} and additional details provided in Appendix~\ref{app:phase_two_training}.

To track and evaluate training dynamics, we build a validation set consisting of trajectories $(v_{1:k}, m_{1:k-1})$. The video clips are disjoint from the training set, and the historical memories are generated by $\pi_0$. We evaluate performance using three metrics: (1) Accuracy, whether the generated memory $m_k$ aligns with the video content; (2) Non-redundancy rate, whether $m_k$ avoids duplicating information already present in the historical memory; (3) Relevance win/tie/loss ratio, how often $m_k$ is more relevant, equally relevant, or less relevant to tasks in environment compared to that from $\pi_0$. Implementation details for these metrics are provided in Appendix~\ref{app:phase_two_training}.

We observe several notable phenomena during phase two training. 

\begin{figure}[th]
    \centering
    \includegraphics[width=0.9\linewidth]{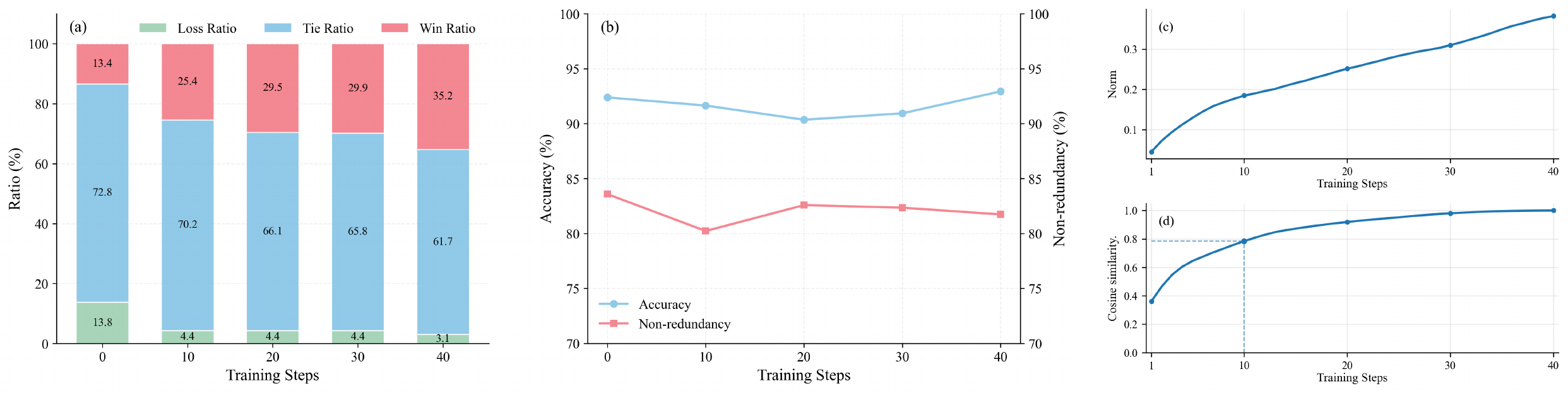}
    \vspace{-5pt}
    \caption{Phase Two training dynamics for the object recognition task on VideoMME, using five questions from the streaming task as environment feedback. (a) training improves task relevance on the validation set; (b) memory quality remains stable; (c) adapter norm increases steadily with training steps;  (d) cosine similarity between the current adapter and the final (40-step) adapter.}
    \label{fig:stage2_training_curve_22}
\end{figure}

\textbf{Training Dynamics.} As shown in Figure~\ref{fig:stage2_training_curve_22}, Phase Two improves task relevance while maintaining stable accuracy and low redundancy.
The adapter direction converges early, with cosine similarity between step-10 and step-40 reaching $0.8$, indicating that later training primarily increases its norm. Motivated by this, we propose a simple acceleration strategy: directly scale the step-10 adapter to a target norm (0.3) to approximate the final adapter. 
As shown in Table~\ref{tab:adapter_scaling}, the scaled adapter achieves performance comparable to step-40, while reducing training data and time by approximately 75\%.





\begin{table}[th]
\centering
\small
\setlength{\tabcolsep}{6pt}
\caption{Comparison of Phase Two training methods against Phase One across quality metrics (accuracy and non-redundancy) and task-relevance metrics (loss/tie/win ratios). Phase Two methods include full-parameter training, 40-step adapter training, and a scaled 10-step adapter with weights multiplied by 1.5. Phase One results are reported as mean $\pm$ standard deviation over three runs.}
\label{tab:adapter_scaling}
\resizebox{0.95\columnwidth}{!}{
\begin{tabular}{lccccc}
\toprule[1pt]
\textbf{Method} & \textbf{Accuracy} $\uparrow$ & \textbf{Non-redundancy} $\uparrow$ & \textbf{Loss Ratio} & \textbf{Tie Ratio} & \textbf{Win Ratio} $\uparrow$ \\
\midrule[0.5pt]
\sysname Phase One & $92.39 \pm 0.47$ & $83.59 \pm 0.43$ & $13.82 \pm 1.15$ & $72.80 \pm 0.83$ & $13.38 \pm 0.37$ \\
w/ full-parameter tuning & 91.29 & 78.21 & 2.00  & 58.00 & 40.00 \\
w/ 40-step adapter & 92.93 & 81.74 & 3.06 & 61.72 & 35.22  \\
w/ scaled adapter & 91.54 & 80.96 & 3.17 & 58.52 & 38.31 \\
\bottomrule[1pt]
\end{tabular}}
\end{table}

\begin{figure}[th]
    \centering
    \includegraphics[width=0.9\linewidth]{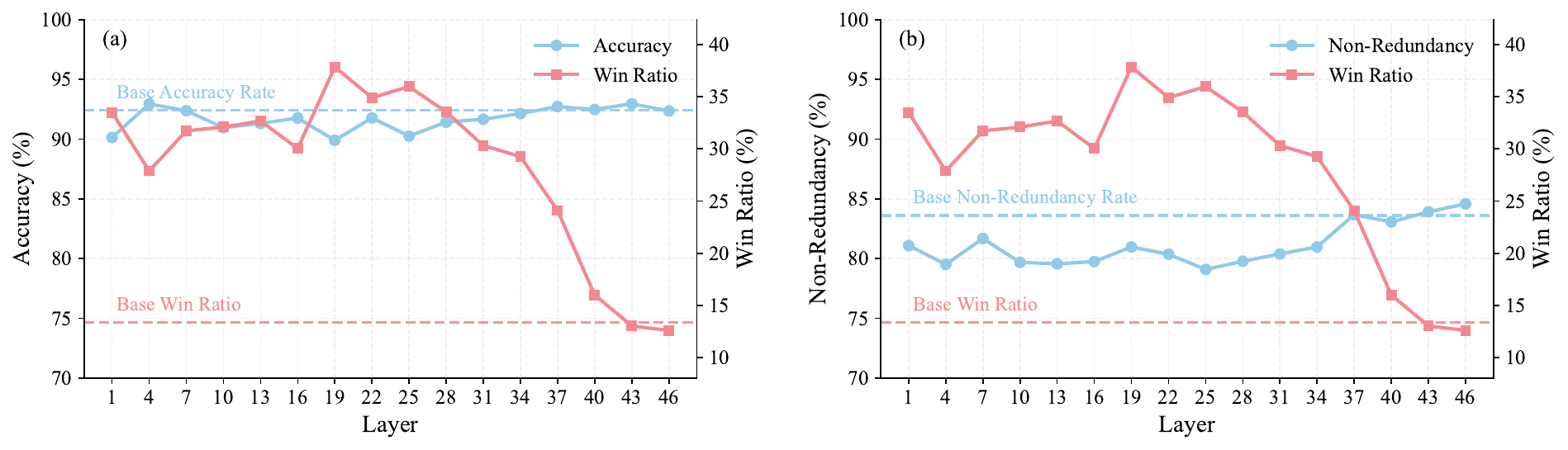}
    \vspace{-5pt}
    \caption{Ablation study on adapter placement across layers on the object recognition task from VideoMME, evaluated by accuracy, non-redundancy rate, and win ratio.}
    \label{fig:layer_ablation_new}
\end{figure}

\textbf{Layer Ablation.} We conduct an ablation study to examine how adapter placement across layers affects performance. Figure~\ref{fig:layer_ablation_new} reports the accuracy, non-redundancy rate, and win ratio for adapters inserted at different layers on the VideoMME object recognition task. The results show that placing adapters in shallow and middle layers is more effective than in deep layers. In all our experiments, we place the adapter at layer 22.




\textbf{Adapter Training Discussion} We motivate our choice of training an adapter vector from three perspectives. First, it is lightweight and parameter-efficient, requiring only 2{,}048 trainable parameters. As shown in Table~\ref{tab:adapter_scaling}, adapter training and full-parameter training achieve comparable win ratios. However, the adapter preserves accuracy and non-redundancy rates close to those of the original model, whereas full-parameter training causes a drop in both metrics. This suggests that adapter training effectively mitigates catastrophic forgetting.

Second, from an inference standpoint, the learned adapter can be viewed as a group of parameters that is loaded only at inference time, leaving the deployed model's weights unchanged. This design preserves deployment efficiency and incurs no additional serving cost. Conceptually, the adapter can also be interpreted as a form of parametric personalized memory.

Third, prior findings show that LLM behaviors can be steered at inference time by adding a vector to the layer's activations, without modifying the model weights~\cite{turner2023steering,panickssery2023steering,arditi2024refusal,chen2025persona}. This works because high-level behaviors are approximately encoded as linear directions in activation space~\cite{elhage2022toy,park2023linear}. Our method follows the same intuition, but rather than extracting such a direction, we learn it directly, so that the resulting vector captures the activation-space representation of each target task.



\subsection{Memory Evaluation Method, Dataset and Metrics} 

We evaluate our approach on three VQA benchmarks:
\begin{itemize}
\item \textbf{VideoMME~\cite{fu2025video}:} The videos in VideoMME are sourced from YouTube and provide comprehensive coverage of diverse video types. Its question categories span twelve tasks, such as object recognition, action reasoning, and counting. Because the memory length of long videos may exceed GPT-4o's maximum token limit, we use only the short and medium subsets of VideoMME. In total, these subsets contain 600 videos and 1,800 QA pairs.

\item \textbf{EgoLife~\cite{yang2025egolife}:} EgoLife consists of egocentric videos depicting practical, everyday activities. Its questions cover five tasks types: event recall, relation map, entity log, habit insight and task master. The dataset contains 500 VQA samples.


\item \textbf{EgoTempo~\cite{plizzari2025omnia}:} EgoTempo is an egocentric VQA benchmark featuring temporal understanding. The benchmark contains 500 VQA samples spanning ten task types, such as locating objects, spatial relationships, and future action prediction. 
\end{itemize}

To simulate a multimodal agent that perceives and processes information sequentially, we reformulate each VQA benchmark into a stream of sequential tasks. For each benchmark, we group video-question pairs by question type, with each group defining a distinct task. This design mimics a specific environment and allows us to assess whether the agent can generate task-relevant memory. Within each task, the agent observes the associated videos one by one and consumes them to generate episodic memory. Each question is posed only after its corresponding video has been seen. 


To test whether \sysname can generate task-relevant memory, the first five questions of each task are answered using memory produced by the Phase One policy. \sysname then performs Phase Two training to update the policy, and the remaining questions are answered using memory generated by the Phase Two policy.

To isolate the evaluation of memory quality, we require each question to be answered using only the generated memory, without access to the original video. Specifically, we use GPT-4o as the answer generator. Given a question and the generated memory, GPT-4o is prompted to first perform reasoning and then determine whether the memory contains sufficient information to answer the question. If not, it returns "insufficient information"; otherwise, it produces an answer. The prompt is provided in Table~\ref{tab:qa_test_prompts}.





We report three complementary metrics to assess memory quality from different angles:
\begin{itemize}
    \item Accuracy: The proportion of all questions that are correctly answered. This reflects the overall utility effectiveness of the memory.
    \item Coverage: The fraction of all questions for which the memory contains the information necessary to answer. This evaluates the comprehensiveness of the memory.
    \item Precision: Among the questions for which the memory is deemed sufficient, the fraction that are answered correctly. This measures the faithfulness of the memory.
\end{itemize}

\subsection{Baselines}

We evaluate \sysname with episodic memories generated by three categories of baselines: (1) Base MLLM Models. This group includes Gemini-1.5-Pro~\cite{team2024gemini}, Gemini-2.5-Pro~\cite{comanici2025gemini}, GPT-5.2~\cite{gpt52}, and Qwen3-VL-30B-A3B~\cite{bai2025qwen3}. We adopt the same streaming generation protocol as \sysname. At each step, the model is provided with the four most recent 10-second video clips and their corresponding memories, along with a new incoming 10-second clip. 
Prompt templates are provided in Table~\ref{tab:prompt_generating_episodic_memory}. (2) Memory Frameworks.
We further compare against recent memory frameworks, including EgoGPT~\cite{yang2025egolife}, HippoMem~\cite{lin2025hippomm}, and M3-Agent~\cite{DBLP:journals/corr/abs-2508-09736}, using their generated episodic memories. 

\textbf{}

\subsection{Main Results}

\begin{table}[t]
\vspace{-30pt}
\caption{Results on VideoMME, EgoLife, and EgoTempo. Best results are in bold, and second-best results are underlined.}
\renewcommand{\arraystretch}{0.88}
\centering
\resizebox{0.95\linewidth}{!}{%
\begin{tabular}{lccccccccc}
\toprule[1pt]
\multicolumn{1}{c}{\multirow{2.5}{*}{\textbf{Method}}} & \multicolumn{3}{c}{\textbf{VideoMME}} & \multicolumn{3}{c}{\textbf{EgoLife}} & \multicolumn{3}{c}{\textbf{EgoTempo}} \\
\cmidrule(l){2-4} \cmidrule(l){5-7} \cmidrule(l){8-10}
& Acc.($\uparrow$) & Cov.  & Prec.($\uparrow$) & Acc.($\uparrow$) & Cov.  & Prec.($\uparrow$) & Acc.($\uparrow$) & Cov.  & Prec.($\uparrow$) \\
\midrule[0.5pt]
EgoGPT & 44.3 & 58.7 & 75.5 & 19.2 & 28.2 & 68.1 & 15.0 & 33.5 & 44.9 \\
HippoMM & 48.9 & 66.6 & 73.5 & 30.4 & 43.4 & 70.0 & 15.8 & 30.8 & 51.1 \\
M3-Agent & 62.5 & 77.7 & 80.4 & 21.8 & 30.8 & 70.8 & 16.0 & 36.3 & 44.2 \\
Gemini-1.5-Pro & 55.3 & 65.9 & 83.9 & 39.4 & 51.6 & 76.4 & 19.7 & 34.3 & 57.4 \\
Gemini-2.5-Pro & 63.2 & 74.8 & \underline{84.4} & \underline{43.8} & 56.6 & \underline{77.4} & 25.8 & 42.3 & 61.0 \\
GPT-5.2 & \underline{67.3} & 80.8 & 83.3 & 34.8 & 48.2 & 72.2 & \textbf{32.1} & 51.4 & \underline{62.4} \\
Qwen3-VL-30B-A3B & 61.6 & 74.7 & 82.5 & 38.4 & 52.4 & 73.3 & 22.3 & 38.9 & 57.2 \\
\sysname & \textbf{67.9} & 79.3 & \textbf{85.6} & \textbf{45.4} & 56.4 & \textbf{80.5} & \underline{27.6} & 43.7 & \textbf{63.2} \\

\bottomrule[1pt]
\end{tabular}
}
\label{tab:main_results}
\vspace{-14pt}
\end{table}

Table~\ref{tab:main_results} presents the performance of \sysname and baseline methods on VideoMME, EgoLife, and EgoTempo, while Table~\ref{tab:videomme}, Table~\ref{tab:egolife}, and Table~\ref{tab:egotempo} further break down results by task across these benchmarks. Compared to Qwen3-VL-30B-A3B, the episodic memory learned in Phase One already improves accuracy and reduces error rates across all benchmarks. Phase Two further refines the memory to better align with questions in the environment, leading to substantial gains in VQA performance, with both accuracy and precision increasing consistently across all benchmarks. Overall, \sysname improves accuracy by 6.3\%, 7.0\%, and 5.3\% on VideoMME, EgoLife, and EgoTempo, respectively. Appendix~\ref{app:robustness} shows these gains persist with a different QA answer generator.

Compared with other baselines, including closed-source models and alternative memory frameworks, \sysname demonstrates strong and consistent performance, outperforming all baselines by a clear margin on both VideoMME and EgoLife. On EgoTempo, \sysname remains competitive, surpassing most methods and only slightly trailing GPT-5.2 in accuracy. This gap is primarily due to the stronger base video understanding capability of GPT-5.2 in describing fine-grained activities, which are central to many EgoTempo questions. Despite this, \sysname achieves higher precision than GPT-5.2, indicating our training leads to more reliable memory with reduced hallucination.

\begin{table}[htbp]
\caption{Ablation study of Phase One and Phase Two training.}
\renewcommand{\arraystretch}{0.88}
\centering
\resizebox{0.95\linewidth}{!}{%
\begin{tabular}{lccccccccc}
\toprule[1pt]
\multicolumn{1}{c}{\multirow{2.5}{*}{\textbf{Method}}} & \multicolumn{3}{c}{\textbf{VideoMME}} & \multicolumn{3}{c}{\textbf{EgoLife}} & \multicolumn{3}{c}{\textbf{EgoTempo}} \\
\cmidrule(l){2-4} \cmidrule(l){5-7} \cmidrule(l){8-10}
& Acc.($\uparrow$) & Cov.  & Prec.($\uparrow$) & Acc.($\uparrow$) & Cov.  & Prec.($\uparrow$) & Acc.($\uparrow$) & Cov.  & Prec.($\uparrow$) \\
\midrule[0.5pt]
Qwen3-VL-30B-A3B & 61.6 & 74.7 & 82.5 & 38.4 & 52.4 & 73.3 & 22.3 & 38.9 & 57.2 \\
+ Task Prompt & 64.2 & {77.3} & 83.0 & \underline{40.0} & 51.0 & \underline{78.4} & \underline{24.5} & 42.9 & 57.1 \\
\sysname Phase One & \underline{64.4} & 75.4 & \underline{85.3} & 39.6 & 51.0 & 77.6 & 23.7 & 37.5 & \textbf{63.3} \\
\sysname & \textbf{67.9} & {79.3} & \textbf{85.6} & \textbf{45.4} & {56.4} & \textbf{80.5} & \textbf{27.6} & {43.7} & \underline{63.2} \\
\bottomrule[1pt]
\end{tabular}
}
\label{tab:ablation}
\vspace{-5pt}
\end{table}

To further validate the benefit of training, we compare \sysname against a prompt-only baseline: at test time, we feed the same recent tasks to Qwen3-VL-30B-A3B and prompt it to generate task-relevant memory (see prompt in Table~\ref{tab:prompt_generating_episodic_memory_with_supplement}). As shown in Table~\ref{tab:ablation}, the prompting approach improves over the base model but still falls short of \sysname, indicating that parameter updates are more effective than prompting alone.
We also ablate the two training phases. Phase One improves both accuracy and precision over the base model, showing it learns a more fundamentally reliable memory. Phase Two yields a further substantial gain in accuracy by generating more task-relevant memory.

\begin{wraptable}{r}{0.48\textwidth}
\vspace{-10pt}
\caption{Adapter analysis on object recognition task.}
\renewcommand{\arraystretch}{0.88}
\centering
\resizebox{1.0\linewidth}{!}{%
\begin{tabular}{lccc}
\toprule[1pt]
\textbf{Method} & Acc.($\uparrow$) & Cov.  & Prec.($\uparrow$) \\
\midrule[0.5pt]
\sysname Phase One & \underline{65.7} & 74.7 & \underline{87.9} \\
w/ Counting adapter & 65.0 & 74.0 & 87.8 \\
w/ OCR adapter & 65.0 & {75.7} & 85.9 \\
w/ Attribute adapter & 65.0 & 74.3 & 87.4 \\
w/ Object adapter & \textbf{69.7} & {78.7} & \textbf{88.6} \\
\bottomrule[1pt]
\end{tabular}
}
\label{tab:ablation_object_recognition}
\vspace{-10pt}
\end{wraptable}

To verify that Phase Two learns task-specific memory focus rather than a generally stronger adapter, we conduct a cross-task transfer test: we fix the evaluation task to Object Recognition VQA and replace its adapter with those trained using feedback from other tasks (Counting, OCR, Attribute Perception). If Phase Two merely produced a universally better adapter, all variants would improve Object Recognition; if it learns task-specific focus, only the matched adapter should help. As shown in Table~\ref{tab:ablation_object_recognition}, only the Object Recognition adapter improves, indicating that Phase Two acquires task-specific focus, rather than a generic capability that transfers across tasks.

\section{Case Study}

\subsection{Task-Focused Memory in Phase Two}

We present several case studies to compare how memory evolves from Phase One to Phase Two. As shown in Case 1 of Table~\ref{table:case_study_phase2}, the memorization policy learned in Phase One produce episodic memory in an accurate, general event-narration style. After further training in Phase Two on questions related to object recognition, the memory style shifts toward capturing fine-grained object details. Table~\ref{table:case_study_phase2} reports three additional cases on different tasks, which together quantitatively demonstrate the effectiveness of our method.

\begin{longtable}{@{}p{\textwidth}@{}}
\caption{Qualitative comparison of episodic memory generations before and after the Phase Two training. Yellow highlights indicate additional fine-grained details generated after training.}
\label{table:case_study_phase2}
\\

\toprule[1pt]
\endfirsthead

\toprule[0.5pt]
\endhead

\midrule[0.5pt]
\multicolumn{1}{r}{\textit{(Continued on next page)}} \\
\endfoot

\bottomrule[1pt]
\endlastfoot

\textbf{Case 1 (Object Recognition)} \\[3pt]

\noindent\textbf{Representative task questions} \\
{\small
\textit{
``Which galaxies are depicted in the video?'';
``Which object does the holder made in this video visually resemble?'';
``Which team scored in the video?'';
``What does the chef in the video end up cutting with a knife?'';
``What is the second class the boys are taking in the video?''
}}
\\[5pt]

\noindent\textbf{Video} (Illustrated as sampled frames) \\[2pt]
\includegraphics[width=\linewidth]{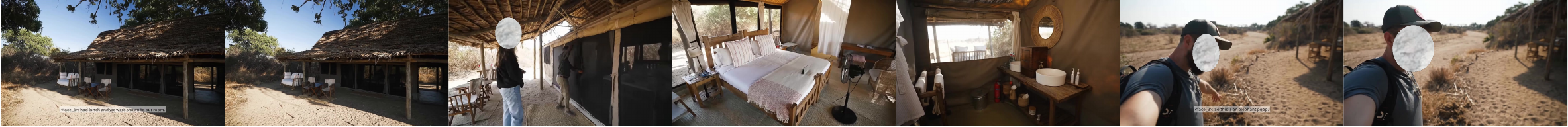}
\\[6pt]

\noindent\textbf{Episodic memory (Phase One)} \\[2pt]
The video transitions to the exterior of a thatched-roof building with a swing and chairs on a sandy area,
followed by a view of a tented room containing a large bed with white linens, a fan, and a bathroom area with a wooden table, sink, and round mirror.
\faceid{3} stands on a dirt path in the arid landscape, pointing at the ground and stating,
``So this is an elephant poop,''
with sparse trees and dry terrain visible in the background.
\\[6pt]

\noindent\textbf{Episodic memory (Phase Two)} \\[2pt]
The video transitions to the exterior of a thatched-roof building with a swing and wooden chairs on a sandy area, surrounded by trees.
\hl{Inside a tented structure, a person adjusts a screen while another stands nearby.}
The tent's interior features a large bed with \hl{white and pink bedding}, a fan, wooden furniture, and
\hl{large windows overlooking the savanna}.
A bathroom area within the tent displays a wooden sink with toiletries, a round mirror, and \hl{towels hanging}.
\faceid{3} then stands on a dirt path, pointing at the ground and stating,
``So this is an elephant poop,''
while \hl{wearing a cap and backpack}, with the \hl{arid landscape and thatched structures} in the background.
\\[10pt]

\midrule[0.5pt]
\\[-2pt]

\textbf{Case 2 (Object Reasoning)} \\[3pt]

\noindent\textbf{Representative task questions} \\
{\small
\textit{
``What is the most likely role of the blonde woman in the video, clad in a blue T-shirt and black shorts?'';
``What is the role of the woman in the video with short hair, wearing a black top, and donning a white scarf?'';
``In which country is the food featured in the video recognized worldwide?'';
``Which player was the winner of the game?'';
``Which of the following elements is not present in the "Starry Sky"?''
}}
\\[5pt]

\noindent\textbf{Video} (Illustrated as sampled frames) \\[2pt]
\includegraphics[width=\linewidth]{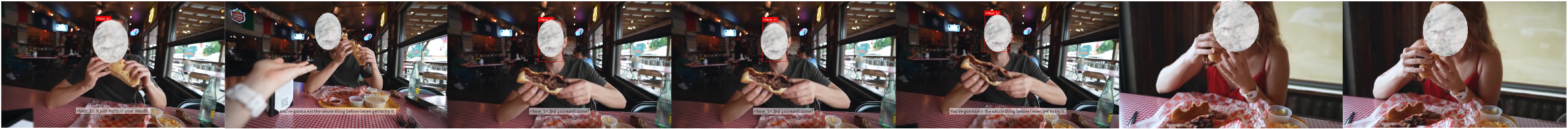}
\\[6pt]

\noindent\textbf{Episodic memory (Phase One)} \\[2pt]
\faceid{1} holds the sandwich, teasing, 
``You're gonna eat the whole thing before I even get to try it.'' 
They grin, adding, 
``Oh, well, we're gonna have to fight to the death for it.'' 
The scene shifts to a person in a red dress sitting at the same table, 
taking a large bite of the hoagie with evident enjoyment, 
their hands gripping the sandwich as they savor the flavors.
\\[6pt]

\noindent\textbf{Episodic memory (Phase Two)} \\[2pt]
\faceid{1} \hl{remains seated at the table}, holding the hoagie and stating, 
``You're gonna eat the whole thing before I even get to try it,'' 
with a playful yet teasing tone. 
Another voice responds, 
``Oh, well, we're gonna have to fight to the death for it,'' 
\hl{adding a lighthearted competitive vibe to the interaction}. 
The scene then shifts to a woman with \hl{long red hair}, wearing a red dress and \hl{a white wristwatch}, 
seated at a similar checkered-tablecloth table. 
She takes a large bite of the hoagie, her expression focused on savoring the flavor, 
with the sandwich's \hl{layers of smoky brisket and jalapeno sausage visible as she eats}. 
The background shows a \hl{warmly lit restaurant interior}, maintaining the casual, communal dining atmosphere.
\\[10pt]

\midrule[0.5pt]
\\[-2pt]

\textbf{Case 3 (Attribute Perception)} \\[3pt]

\noindent\textbf{Representative task questions} \\
{\small
\textit{
``From which country does the first-place athlete in the video originate?'';
``What color are the foxes in the video?'';
``Which genre best describes the performance based on the activity and costumes?'';
``What is the shape of the paper shown in the video?'';
``What can be inferred about the tone of the play?''
}}
\\[5pt]

\noindent\textbf{Video} (Illustrated as sampled frames) \\[2pt]
\includegraphics[width=\linewidth]{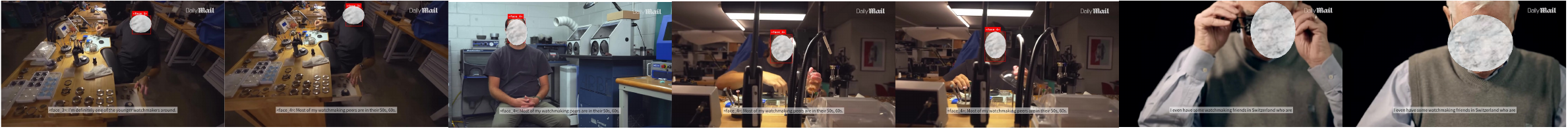}
\\[6pt]

\noindent\textbf{Episodic memory (Phase One)} \\[2pt]
\faceid{4} sits in the workshop, stating, 
``Most of my watchmaking peers are in their 50s, 60s.'' 
The scene then cuts to an older man adjusting a magnifying glass against a black background, saying, 
``I even have some watchmaking friends in Switzerland who are''.
\\[6pt]

\noindent\textbf{Episodic memory (Phase Two)} \\[2pt]
\faceid{4} sits in a workshop \hl{surrounded by machinery, 
wearing a dark shirt and khaki pants, with hands clasped} as they speak: 
``Most of my watchmaking peers are in their 50s, 60s.'' 
The scene transitions to an older man with \hl{white hair}, 
dressed in a \hl{light blue shirt and gray vest}, 
adjusting a magnifying glass while stating, 
``I even have some watchmaking friends in Switzerland who are''. 
The older man's \hl{focused expression and deliberate hand movements} 
emphasize the precision of watchmaking craftsmanship.
\\[10pt]

\midrule[0.5pt]
\\[-2pt]

\textbf{Case 4 (Counting Problem)} \\[3pt]

\noindent\textbf{Representative task questions} \\
{\small
\textit{
``How many red flags appear in the video?'';
``WHow many outfits did the male protagonist change in total in the video?'';
``What is the total number of bird species that are visible in the video?'';
``How many tricks are performed in this video?'';
``How many people join in a toast at the end of the video?''
}}
\\[5pt]

\noindent\textbf{Video} (Illustrated as sampled frames) \\[2pt]
\includegraphics[width=\linewidth]{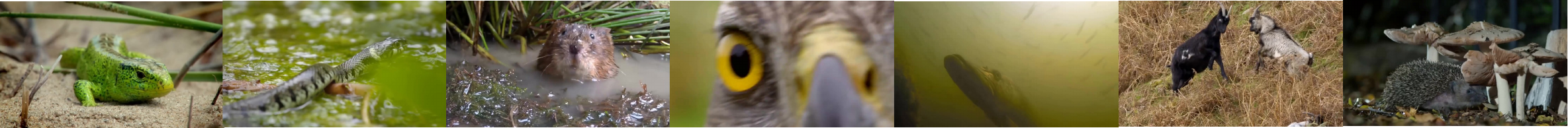}
\\[6pt]

\noindent\textbf{Episodic memory (Phase One)} \\[2pt]
A vibrant green lizard with a yellow head rests on sandy ground, its patterned scales blending with the surroundings. 
A snake glides through greenish water, partially submerged among aquatic plants. 
A beaver emerges from murky water, navigating reeds and vegetation with wet fur glistening. 
The intense yellow eye of an owl fills the frame, feathers ruffled. 
A fish swims in murky water, its form barely visible. 
Two goats clash horns in a field of dry, golden grass, their movements vigorous. 
Finally, a young otter lounges on seaweed, damp fur resting near the shoreline.
\\[6pt]

\noindent\textbf{Episodic memory (Phase Two)} \\[2pt]
A vibrant green lizard with intricate patterns rests on sandy ground, its head slightly raised as it observes its surroundings amid thin green stems and dry grass. 
The scene shifts to a snake with a \hl{patterned body} gliding through murky water, its head above the surface among aquatic plants. 
\hl{A water vole with brown fur and prominent whiskers emerges from murky water, surrounded by aquatic vegetation.} 
A close-up reveals an owl's striking yellow eye with detailed feather patterns, capturing its intense gaze. 
Underwater, a fish swims in yellowish water with other fish in the background. 
Two goats---\hl{one black, one white}---clash horns in tall dry grass, their bodies tense during the struggle. 
A young otter \hl{with wet brown fur} lies on seaweed-covered rocks, eyes wide and alert. 
Finally, \hl{a hedgehog with spiny brown fur sniffs around large mushrooms in a forest, surrounded by fallen leaves and dark foliage}.
\\[10pt]

\end{longtable}



\subsection{A Complete Episodic Memory Example on a Long Video}

In this subsection, we present a complete episodic memory generated by \sysname for a five-minute long video, 
shown in Table~\ref{tab:case_coherent_memory_superman}, to directly demonstrate the key advantages of our memory: (1) Detailed yet low-hallucination content. For this five-minute video, the memorization policy generated highly detailed yet accurate descriptions, resulting in a memory of 4620 tokens. (2) Coherent streaming generation. Our framework yields episodic memory with smooth narrative flow across clip boundaries. Moreover, we maintain a global entity ID for each key entity (e.g., \facebox{1}), ensuring consistent entity references throughout the long-form description. (3) Task relevance. This example is produced under the counting task. The memorization policy emphasizes counting related-information. For instance, it records important local counts within individual clips, which helps to correctly solve the global counting problem, a representative and challenging long-horizon problem.

\newcommand{\ansevidence}[1]{\textcolor{blue}{\textbf{Answer Evidence:} #1}}

\begin{longtable}{@{}p{\textwidth}@{}}
\caption{A complete episodic memory example on a long video. Square-bracketed tokens (e.g. \facebox{1}) are used for entity tracking. The question is shown before the generated memory to demonstrate that the memory preserves sufficient task-relevant information to answer the question correctly. The example shows that the generated memory is accurate, coherent, and task-relevant. \textcolor{blue}{Task-relevant memories are highlighted in blue.} The video and question are from VideoMME, and the original video is available at: \url{https://www.youtube.com/watch?v=uC9qU3X1JgM}. 
}
\label{tab:case_coherent_memory_superman}
\\

\toprule[1pt]
\endfirsthead

\toprule[0.5pt]
\endhead

\midrule[0.5pt]
\multicolumn{1}{r}{\textit{(Continued on next page)}} \\
\endfoot

\bottomrule[1pt]
\endlastfoot

\textbf{Question.} 
How many persons does Superman fight versus? \\[2pt]

\textbf{Options.} 
A. 6. \quad
B. 4. \quad
C. 5. \quad
D. 7. \\[2pt]

\textbf{Ground-truth Answer.} 
C. 5. \\[2pt]

\textbf{Model Prediction.} 
C. 5. \\[4pt]

\textbf{Generated Memory.} \\[2pt]

\textbf{00:00:00--00:00:10} \\
A shirtless man with dark hair stands with his back to the camera, facing a group of superheroes at a memorial site featuring a reflecting pool, scattered flowers, and city buildings under twilight. 
The group includes \facebox{1} in a red suit, \facebox{2} in a dark armored suit, and \facebox{3} (Wonder Woman) in golden-blue armor with a tiara, sword, and shield. 
\facebox{3} turns to face the shirtless man and speaks: ``He's back.'' 
The scene includes police cars and onlookers in the background, emphasizing a moment of tension or reunion. \\[3pt]

\textbf{00:00:10--00:00:20} \\
The shirtless man identified as \facebox{4} stands with a serious expression, facing the group of superheroes at the memorial site. 
\facebox{2} (Aquaman) holds a trident, clad in his armored suit, while \facebox{3} (Wonder Woman) remains in her golden-blue armor with a sword and shield. 
\facebox{1} (The Flash) in his red suit and \facebox{5} (Cyborg) with a glowing chest stand alongside them. 
The reflecting pool, scattered flowers, and cityscape under twilight persist in the background, maintaining the scene's tense atmosphere as the group collectively observes \facebox{4} \\[3pt]

\textbf{00:00:20--00:00:30} \\
\facebox{3} (Wonder Woman) in golden-blue armor with a tiara, sword, and shield turns to face \facebox{4}, stating, ``He's back.'' 
Her expression is intense. 
\facebox{4}, shirtless with dark hair, stands with a stern look toward the group. 
The scene cuts to \facebox{2} (Aquaman) in his armored suit, long hair and beard framing a serious expression, followed by \facebox{1} (The Flash) in his red suit with a lightning bolt emblem, also solemn. 
\facebox{4} remains tense, then the group appears in a blue-lit environment where their muscular forms are briefly highlighted, suggesting a power-related moment, while the memorial's reflecting pool and cityscape persist in the background. \\[3pt]

\textbf{00:00:30--00:00:40} \\
A close-up of \facebox{4}'s intense eyes fills the frame, emphasizing tension. 
The scene shifts to \facebox{5} (Cyborg) with a glowing chest, stating, ``Scanning him.'' 
\facebox{1} (The Flash) in his red suit reacts with, ``What?'' as \facebox{5} continues, ``Arthur, you need to relax. Your adrenaline's spiking,'' while the cityscape and memorial site remain in the background, maintaining the scene's urgent atmosphere. \\[3pt]

\textbf{00:00:40--00:00:50} \\
The shirtless man \facebox{4} stands on the memorial steps, facing Wonder Woman and The Flash. 
He asks, ``Or show our bellies?'' with a tense expression. 
Cyborg \facebox{5} intervenes, saying ``Victor, stop'' while raising a hand in a restraining gesture. 
Wonder Woman and The Flash remain alert, their postures tense as they respond to the interaction. 
The memorial's reflecting pool, cityscape, and scattered flowers persist in the background, maintaining the scene's urgent tone. \\[3pt]

\textbf{00:00:50--00:01:00} \\
\facebox{5} (Cyborg) with a glowing chest and metallic armor speaks urgently, saying, ``Victor, it's my armor's defense system. It's stronger since the interface,'' while raising a defensive hand. 
\facebox{3} (Wonder Woman) intervenes, shouting, ``Victor, no!'' as \facebox{4} (shirtless man on the memorial steps) adds, ``You're gonna cause an alarm,'' his expression tense. 
The Flash (\facebox{1}) and Aquaman (\facebox{2}) stand alert nearby, their postures rigid, while the cityscape and memorial pool remain in the background, heightening the scene's urgency. \\[3pt]

\textbf{00:01:00--00:01:10} \\
\textcolor{blue}{The shirtless man \facebox{4} places his hand on his pants as Cyborg \facebox{5} fires a red energy beam toward him, causing an explosion on the memorial steps.} 
Smoke billows as an aerial view reveals the chaos, with the group scattered and the memorial site in disarray. 
Wonder Woman \facebox{3} and The Flash \facebox{1} react with shock, while Aquaman \facebox{2} looks on in concern. 
The cityscape and reflecting pool remain visible, amplifying the scene's tension as the aftermath of the blast unfolds. \\[3pt]

\textbf{00:01:10--00:01:20} \\
Wonder Woman \facebox{3} shouts ``Kal-El, no!'' as Cyborg \facebox{5} fires an intense red energy beam, triggering a massive explosion that illuminates the memorial site. 
The Flash \facebox{1} and Aquaman \facebox{2} brace themselves against the blast's force, while the shirtless man \facebox{4} stands amid rising smoke and debris. 
Flames and light fill the frame as the explosion's shockwave disrupts the surrounding area, with the city skyline and reflecting pool now partially obscured by the chaos. \\[3pt]

\textbf{00:01:20--00:01:30} \\
The Flash \facebox{1} rises from the ground, expression bewildered, as he mutters, ``He's confused.'' 
Wonder Woman \facebox{3} exclaims, ``He doesn't know who he is,'' while scanning the smoldering memorial site. 
The Flash adds, ``That cemetery,'' his voice tense. 
Meanwhile, the shirtless man \facebox{4} hoists a massive stone from the steps, muscles straining, as Aquaman \facebox{2} and Wonder Woman ready their weapons, bracing for further conflict amid the lingering smoke and scattered debris. \\[3pt]

\textbf{00:01:30--00:01:40} \\
\textcolor{blue}{Wonder Woman \facebox{3} sprints toward the conflict, her golden armor gleaming as she calls out, ``Arthur, we need to restrain him.'' 
Aquaman \facebox{2} follows, trident raised, while the shirtless man \facebox{4} struggles against their combined effort. 
The Flash \facebox{1} dashes in, red suit flashing, as Wonder Woman and Aquaman grapple with the muscular figure.} 
Smoke from the earlier explosion lingers in the air, mixing with the city's glow as the trio maneuvers around the memorial steps, their movements urgent and coordinated in the chaos. \\[3pt]

\textbf{00:01:40--00:01:50} \\
\textcolor{blue}{Wonder Woman \facebox{3} lies on the memorial steps, her shield scattered nearby, as she regains her footing. 
She rises swiftly, lasso in hand, as the shirtless man \facebox{4} unleashes a golden energy beam toward her. 
The Flash \facebox{1} zips in to intercept, creating a streak of red, while Aquaman \facebox{2} charges with his trident. 
The city's glow illuminates the scene as smoke from the earlier blast lingers. 
Wonder Woman shouts, ``You got it!'' as she dodges the beam, her golden armor reflecting the light. 
The shirtless man \facebox{4} grins, his muscles taut, as he continues the assault, the energy beam slicing through the air. 
Aquaman and The Flash work to flank him, but the muscular figure moves with impossible speed, his bare chest glistening under the city lights.} \\[3pt]

\textbf{00:01:50--00:02:00} \\
\facebox{4} maintains his stance, emitting a powerful golden energy beam from both hands as he targets \facebox{3}. 
\facebox{3} grips the glowing lasso, twisting her body to deflect the beam while shouting, ``Kal-El?'' 
The beam cuts through the air, but she uses the lasso to redirect its force. 
\facebox{1} rushes in as a red blur, attempting to intercept \facebox{4}, while \facebox{2} charges forward with his trident, aiming to flank him. 
Smoke from the earlier explosion lingers, mixing with the city's glow as debris scatters across the memorial steps. 
\facebox{4} grins intensely, his muscles straining under the effort, undeterred by the coordinated attacks from Wonder Woman and Aquaman. \\[3pt]

\textbf{00:02:00--00:02:10} \\
\facebox{3} grips the glowing lasso, her expression intense as she declares, ``Kal-El, the last son of Krypton.'' 
She pulls the lasso taut, redirecting the golden energy beam while urging, ``Remember who you are.'' 
\facebox{4} stands firm, muscles rippling under the city lights, his gaze locked on \facebox{3} as he responds with a strained voice. 
The memorial site chaos continues, debris scattering as \facebox{1} (The Flash) zips past in a red blur, attempting to aid \facebox{3}. 
Smoke from previous explosions mingles with the urban glow, highlighting the tension between the characters. 
\facebox{3} presses forward, lasso glowing brighter, as she demands, ``Tell me who you''---her voice cutting through the din of battle. \\[3pt]

\textbf{00:02:10--00:02:20} \\
Wonder Woman \facebox{3} and Aquaman \facebox{2} maintain a firm grip on \facebox{4}, their combined strength overpowering his resistance as he struggles against their restraint. 
The Flash \facebox{1} zips in, adding his speed to secure \facebox{4}, whose golden energy beam flickers and fades. 
Wonder Woman, lasso still glowing, repeats, ``Remember who you are,'' her voice cutting through the chaos. 
Smoke from earlier explosions drifts over the memorial steps, where debris litters the ground. 
Aquaman's trident glints as he holds \facebox{4} steady, while Wonder Woman's expression shifts from urgency to determination as they finally subdue him. \\[3pt]

\textbf{00:02:20--00:02:30} \\
Wonder Woman \facebox{3} keeps the glowing lasso taut around \facebox{4}, her expression intense as she leans close. 
\facebox{4} trembles, his golden energy fading, while Aquaman \facebox{2} holds him firm with a trident. 
The Flash \facebox{1} stands nearby, red suit crackling with electricity. 
Wonder Woman murmurs, ``You are Kal-El, the last son of Krypton,'' as \facebox{4}'s eyes flicker with recognition, his muscular frame slackening under their restraint. 
Debris and smoke linger on the memorial steps, city lights reflecting off the tension between them. \\[3pt]

\textbf{00:02:30--00:02:40} \\
\facebox{1} (The Flash) bursts into the scene, red suit crackling with electric energy as he zips toward \facebox{4}. 
Lightning trails behind him, casting sharp shadows on the memorial steps. 
He unleashes a pulse of blue energy, disrupting \facebox{4}'s fading golden aura. 
\facebox{4} wobbles, muscles straining as \facebox{3} (Wonder Woman) tightens the glowing lasso around his wrist and \facebox{2} (Aquaman) anchors him with a trident. 
Wonder Woman's voice rings out: ``You're not alone anymore.'' 
The Flash steps back, breathless, as \facebox{4}'s eyes dim, his body collapsing under their combined hold. 
Debris scatters across the wet pavement, city lights reflecting off the tension as the trio secures \facebox{4}, ensuring he cannot resist further. \\[3pt]

\textbf{00:02:40--00:02:50} \\
The trio stands amidst the debris of the memorial site, city lights casting a blue glow over the scene. 
\facebox{3} (Wonder Woman) releases the lasso, her hand resting on \facebox{4}'s shoulder as he breathes heavily, his golden aura gone. 
\facebox{2} (Aquaman) lowers his trident, while \facebox{1} (The Flash) gestures toward the skyline, saying, ``We need to move. The city's in chaos.'' 
They begin walking away, the city's glow reflecting off their determined expressions as smoke lingers in the background. \\[3pt]

\textbf{00:02:50--00:03:00} \\
\facebox{4} and \facebox{1} engage in a fierce battle, with \facebox{4} delivering heavy punches while \facebox{1} evades using lightning-fast movements, electric arcs flaring around his red suit. 
\facebox{1} counters with a blast of blue energy, forcing \facebox{4} to stagger as debris scatters across the memorial steps. 
Wonder Woman \facebox{3} and Aquaman \facebox{2} watch from the sidelines, weapons poised, as \facebox{4} attempts a spinning kick that \facebox{1} sidesteps, retaliating with a rapid strike that sends \facebox{4} crashing into a shattered stone pillar. 
The city skyline glows behind them, reflecting off the tension in their clash as lightning continues to crackle around \facebox{1}'s form. \\[3pt]

\textbf{00:03:00--00:03:10} \\
\facebox{19} stands in a cityscape of towering skyscrapers, smoke rising from a burning vehicle and a police car in the background. 
Dressed in a dark, armored suit with a cape, \facebox{19} surveys the chaos, his cowl shadowing his face. 
Nearby, \facebox{4} appears shirtless, muscles taut and skin glistening with sweat, as city lights reflect off his form. 
\textcolor{blue}{\facebox{19} shifts slightly, assessing the scene amid scattered debris and flickering urban lights, while \facebox{4} maintains a tense posture, the aftermath of the earlier battle lingering in the air.} \\[3pt]

\textbf{00:03:10--00:03:20} \\
\facebox{19} (Batman) stands amidst smoldering wreckage and flickering city lights, his armored suit glistening under the evening sky. 
He faces \facebox{4} (Clark), who is shirtless, muscles taut, and sweat glistening on his skin. 
\facebox{4} meets \facebox{19}'s gaze and says, ``I know you,'' his voice steady amid the chaos. 
Behind them, burning vehicles and scattered debris mark the battle's aftermath, while Wonder Woman approaches with her lasso coiled and shield raised. 
The city's glow reflects off the wet pavement, highlighting the tension between the two figures as they stand locked in a silent understanding. \\[3pt]

\textbf{00:03:20--00:03:30} \\
\facebox{3} (Wonder Woman) stands near a reflective water feature, her golden tiara gleaming under the city's twilight glow. 
She declares, ``We do this,'' as she grips her lasso and shield, her expression resolute. 
Facing \facebox{4} (shirtless, sweat glistening on his muscles), she launches a swift punch, prompting \facebox{4} to evade with an agile sidestep. 
The two clash in a flurry of movements---Wonder Woman's lasso whips through the air, while \facebox{4} counters with rapid strikes, debris scattering as their battle intensifies against the backdrop of burning vehicles and towering skyscrapers. \\[3pt]

\textbf{00:03:30--00:03:40} \\
\facebox{3} lies on the wet pavement, her golden tiara askew and shield resting beside her, the aftermath of the clash with \facebox{4} evident in her strained posture. 
\facebox{4} stands over her, chest heaving, sweat dripping down his muscular frame as he glances toward \facebox{19} (Batman), who approaches from the smoldering wreckage. 
\facebox{19}'s armored suit gleams under city lights, his cowl shadowing his face as he surveys the scene. 
The ``METROPOLIS'' police car is visible in the background, its lights pulsing amid smoke from burning vehicles. 
\facebox{4} maintains a tense stance, ready to react as \facebox{19} closes the distance, the city skyline reflecting off the rain-slicked ground. \\[3pt]

\textbf{00:03:40--00:03:50} \\
\facebox{19} crouches beside the ``METROPOLIS'' police car, smoke rising from nearby wreckage. 
He speaks urgently, ``Alfred, I need the big guns,'' as \facebox{4} steps forward. 
\facebox{4} grabs \facebox{19}'s cowl with both hands, pulling it taut before lifting \facebox{19} off the ground. 
The city skyline looms behind them, skyscrapers reflecting the chaos as \facebox{4} holds \facebox{19} aloft, muscles straining under the evening light. \\[3pt]

\textbf{00:03:50--00:04:00} \\
\facebox{4} grips \facebox{19}'s cowl tightly, his voice steady as he says, ``You knew this.'' 
\facebox{19} responds, ``I had to,'' while struggling against the hold. 
\facebox{4} leans in, intensity in his eyes, declaring, ``You won't let me live.'' 
\facebox{19} meets his gaze, replying, ``The world needs you.'' 
As the tension crackles, \facebox{4} releases \facebox{19}, who stumbles back, adjusting his cowl. 
The city skyline looms behind them, skyscrapers reflecting the chaos of burning vehicles and flickering lights, while the two stand locked in a moment of unresolved conflict. \\[3pt]

\textbf{00:04:00--00:04:10} \\
\facebox{4} stands with a tense expression, his voice firm as he declares, ``It doesn't need you.'' 
\facebox{19} remains in the grip of the moment, one hand still near his cowl as he processes the words, the city's chaos unfolding behind them with smoke rising from distant fires and skyscrapers towering into the twilight. \\[3pt]

\textbf{00:04:10--00:04:20} \\
\facebox{4} grips \facebox{19}'s cowl firmly, leaning in with a challenging expression as he asks, ``Do you bleed?'' 
\facebox{19} remains tense, one hand still near his cowl while the mask shifts slightly to reveal a portion of his face, eyes narrowed in response. 
The city skyline, dotted with illuminated skyscrapers and lingering smoke from earlier clashes, frames the confrontation. 
\facebox{19} does not speak immediately, the weight of the question hanging in the air as the two figures stand locked in a silent battle of wills. \\[3pt]

\textbf{00:04:20--00:04:30} \\
Inside a luxury car, \facebox{21} sits in the backseat wearing a brown leather jacket and glasses, the ``MAYBACH'' emblem visible on the door. 
The scene shifts to \facebox{4}, shirtless with dark pants, standing against the city skyline as he engages in a tense confrontation. 
Meanwhile, \facebox{22}---a woman with long red hair in a dark coat over a blue shirt---runs across the grassy area toward a ``METROPOLIS'' police car marked ``8202,'' with an officer holding a weapon behind her. 
Her expression is urgent as she approaches the vehicle, the backdrop of illuminated skyscrapers and lingering smoke from earlier conflicts underscoring the chaotic atmosphere. \\[3pt]

\textbf{00:04:30--00:04:40} \\
\facebox{22} stands beside the ``METROPOLIS'' police car marked ``8202,'' her posture rigid with urgency as she faces \facebox{4}, who approaches from the grassy expanse. 
The city skyline, dotted with glowing skyscrapers and wisps of smoke from earlier chaos, looms behind them. 
\facebox{4}, shirtless and clad in dark pants, moves with deliberate purpose, his expression tense as the unresolved tension from their earlier exchange hangs heavy in the air. 
The officer with the weapon remains in the background, adding to the scene's charged atmosphere. \\[3pt]

\textbf{00:04:40--00:04:50} \\
\facebox{22} faces \facebox{4} with a pleading expression, her voice trembling as she says, ``Please.'' 
She stands close to the ``METROPOLIS'' police car, her dark coat slightly disheveled from running. 
\facebox{4} remains rigid, his shirtless torso glistening under the city lights as he stares back, the tension between them palpable. 
The backdrop of towering skyscrapers, glowing windows, and lingering smoke from earlier chaos underscores the urgency of the moment, while the officer with the weapon stays positioned behind the police car, adding to the scene's intensity. \\[3pt]

\textbf{00:04:50--00:05:00} \\
\facebox{4} stands close to \facebox{22}, his shirtless torso glistening under the city lights as he meets her pleading gaze. 
\facebox{22}'s hand brushes against his shoulder, tears tracing her cheeks as she whispers, ``Please,'' her voice trembling against the backdrop of the ``METROPOLIS'' police car and towering skyscrapers shrouded in smoke. \\

\end{longtable}

\section{Related Work}
\subsection{Long-term Memory in Multi-modal Agent}

Multi-modal agents require external memory to preserve information beyond the context window \cite{hu2025memory, DBLP:journals/corr/abs-2602-06052}. Prior work has explored a broad range of memory paradigms for long-horizon multi-modal reasoning, including memory banks and sparse memory representations for long-video understanding \cite{DBLP:conf/cvpr/0004LJJCSSL24, DBLP:conf/cvpr/SongCWZZWCG0ZLH24}, structured and heterogeneous memories built from textual, object-centric, episodic, semantic, and visual representations \cite{DBLP:conf/eccv/FanMWDLGL24, DBLP:journals/corr/abs-2508-09736,DBLP:journals/corr/abs-2512-02425}, and more advanced memory systems with continual consolidation, structured retrieval, or neuro-symbolic reasoning \cite{DBLP:journals/corr/abs-2601-06037, DBLP:journals/corr/abs-2512-03627, jiang2026advancing}. Recent studies also extend multi-modal memory to personalized, verifiable, and benchmarked agent settings \cite{DBLP:journals/corr/abs-2602-07624, DBLP:journals/corr/abs-2602-00415, DBLP:journals/corr/abs-2601-03515}. As multi-modal observations accumulate over time, maintaining memory quality becomes increasingly challenging, which may negatively affect downstream retrieval and reasoning \cite{DBLP:journals/corr/abs-2505-16067,lu2026mma,DBLP:conf/emnlp/SalamaCYCSZB25}. 

Current approaches typically predefine what is stored in memory, either through prompting~\cite{DBLP:conf/eccv/FanMWDLGL24,DBLP:journals/corr/abs-2601-06037,jiang2026advancing} or via post-training~\cite{DBLP:journals/corr/abs-2508-09736,DBLP:journals/corr/abs-2512-03627}. However, they overlook a critical aspect: what to memorize should dynamically adapt to the environment. An effective multimodal agent should leverage environmental feedback to continuously refine memory formation to ensure not only accuracy but also task relevance. Enabling such continual adaptation is the central focus of our work.

\subsection{Test-Time Training}
Test-time training (TTT) adapts model parameters during inference using data related to the current test instance, enabling better alignment with the current task distribution. The idea traces back to early studies on local learning~\cite{bottou1992local}. More recently, TTT has been explored in large language models. \cite{hardttest} adapts models during inference time using retrieved nearest neighbors from the training corpus and \cite{hubotterefficiently} further improve this paradigm by actively selecting data. TTT is closely related to continual learning~\cite{kirkpatrick2017overcoming,hung2019compacting,douillard2022dytox}, where models are incrementally updated from a stream of data~\cite{chenadaptive}. A central challenge is catastrophic forgetting~\cite{kirkpatrick2017overcoming,french1999catastrophic,hutest}, where adapting to new data may degrade previously acquired knowledge. 

Existing TTT frameworks typically rely on direct adaptation signals, such as task inputs or self-supervised objectives derived from the test instances~\cite{hardttest,hutest,akyurek2025surprising}. In contrast, our setting involves indirect feedback. During deployment, a multimodal agent cannot directly observe signals reflecting memory quality, instead it only interacts with tasks in the environment. This creates a gap between observable experience and the optimization objective for memory.


\section{Conclusion}

In this paper, we introduce \sysname, a reinforcement learning framework that trains a memorization policy to generate task-relevant memory. The framework follows a two-phase design: Phase One optimizes the memorization policy with multi-objective rewards to produce accurate, non-redundant, well-formatted, and content-rich episodic memories. Phase Two further aligns the policy toward more task-relevant content through tuning a lightweight adapter on the base MLLM. We evaluate \sysname on VideoMME, EgoLife, and EgoTempo under a streaming VQA setting. Across all three benchmarks, \sysname consistently outperforms all baselines, including closed-source MLLMs and existing memory frameworks, with Phase One enhancing fundamental memory quality and Phase Two further aligning memory content with task demands. Future work will extend \sysname beyond episodic memory to semantic and visual memory, and explore adaptive memorization in more interactive embodied environments.

\clearpage

\bibliographystyle{plainnat}
\bibliography{main}

\clearpage

\beginappendix

\section{Implementation Details of Tools}\label{appendix:tools}

Here, we provide the implementation details of the tools for representation extraction introduced in Section~\ref{subsection:problem_formulation}.

\noindent\textbf{Facial Recognition} For face recognition, we follow the previous work~\cite{DBLP:journals/corr/abs-2508-09736} to obtain bounding boxes and filter low-quality faces. Then, we draw the detected bounding boxes in the video. For each 1-second clip in the video, we select the middle frame, draw bounding boxes around all detected faces, and extend each bounding box forward and backward in time by 0.5 seconds respectively. We also annotate the corresponding face ID in the upper-left corner of each bounding box.

\noindent\textbf{Voice Identification} For speech recognition, we use Gemini-2.5-Pro to extract audio segments and identify speaker. 
Speaker identification is performed based on the current clip. The model matches the recognized audio segments with the face ID of their most likely corresponding speaker. Meanwhile, for segments where the speaker cannot be identified, the model retrieves similar voices from the global memory and labels them with the corresponding face ID. Finally, the speech content, together with the speaker's face ID, is displayed as subtitles at the bottom of the video.
The prompt used for speech recognition and speaker identification are shown in Table~\ref{tab:prompt_voice_identification} and Table~\ref{tab:prompt_speaker_identification}.

\begin{longtable}{P}
\toprule[1pt]
\textbf{The Prompt for Speech Recognition}\\
\midrule[0.5pt]

\endfirsthead

\toprule[0.5pt]
\endhead

\midrule[0.5pt]
\multicolumn{1}{r}{\textit{(Continued on next page)}}
\endfoot

\endlastfoot

You are given a video with a total duration of 10 seconds. Your task is to perform Automatic Speech Recognition (ASR) and audio diarization on the provided video. Extract all speech segments with accurate timestamps and segment them by speaker turns (i.e., different speakers should have separate segments), but without assigning speaker identifiers. \\
 \\
Return a JSON list where each entry represents a speech segment with the following fields: \\
- start\_time: Start time in seconds, represented as a floating-point number, accurate to 0.1s. \\
- end\_time: End time in seconds, represented as a floating-point number, accurate to 0.1s. \\
- asr: The transcribed text for that segment. \\
 \\
Example Output: \\
\textasciigrave \textasciigrave \textasciigrave json \\ {}
[ \\
\qquad \{\textquotedbl start\_time\textquotedbl : 5.3, \textquotedbl end\_time\textquotedbl : 6.9, \textquotedbl asr\textquotedbl : \textquotedbl Hello, everyone.\textquotedbl \}, \\
\qquad \{\textquotedbl start\_time\textquotedbl : 9.2, \textquotedbl end\_time\textquotedbl : 11.6, \textquotedbl asr\textquotedbl : \textquotedbl Welcome to the meeting.\textquotedbl \} \\
] \\

\textasciigrave \textasciigrave \textasciigrave  \\
 \\
Strict Requirements: \\
 \\
- Ensure precise speech segmentation with accurate timestamps. \\
- Adjacent sentences need to be separated. Each list item can only have one sentence. \\
- Preserve punctuation and capitalization in the ASR output. \\
- Skip the speeches that can hardly be clearly recognized or extremely SHORT in time. \\
- Return only the valid JSON list (which starts with \textquotedbl [\textquotedbl  and ends with \textquotedbl ]\textquotedbl ) without additional explanations. \\
- If the video contains no speech, return an empty list (\textquotedbl []\textquotedbl ). \\
	 \\
Now generate the JSON list based on the given video: \\

\bottomrule[1pt]
\caption{Prompt used for automatic speech recognition.}
\label{tab:prompt_voice_identification}
\end{longtable}

\begin{longtable}{P}
\toprule[1pt]
\textbf{The Prompt for Speaker Identification.}\\
\midrule[0.5pt]

\endfirsthead

\toprule[0.5pt]
\endhead

\midrule[0.5pt]
\multicolumn{1}{r}{\textit{(Continued on next page)}}
\endfoot

\endlastfoot

You are given a video. Your task is to match the subtitle with the <face\_id> of its speaker. \\
The subtitle to be matched is given in the following JSON list: \\
\textasciigrave \textasciigrave \textasciigrave json \\
\{subtitles\} \\
\textasciigrave \textasciigrave \textasciigrave  \\
 \\
The returned list must have the same length as the input JSON list.  \\
Each item in the list shall include an additional string field named \textquotedbl speaker\textquotedbl , with the value determined as follows: \\
- If the corresponding subtitle is definitively associated with a <face\_id>, set \textquotedbl speaker\textquotedbl ~to that <face\_id>; \\
- Otherwise, set \textquotedbl speaker\textquotedbl ~to the string literal \textquotedbl unknown\textquotedbl . \\
 \\
Example Output: \\ 
\textasciigrave \textasciigrave \textasciigrave json \\ {}
[ \\
\qquad \{\textquotedbl start\_time\textquotedbl : 5.3, \textquotedbl end\_time\textquotedbl : 6.9, \textquotedbl asr\textquotedbl : \textquotedbl Hello, everyone.\textquotedbl , \textquotedbl speaker\textquotedbl : \textquotedbl <face\_1>\textquotedbl \}, \\
\qquad \{\textquotedbl start\_time\textquotedbl : 9.2, \textquotedbl end\_time\textquotedbl : 11.6, \textquotedbl asr\textquotedbl : \textquotedbl Welcome to the meeting.\textquotedbl , \textquotedbl speaker\textquotedbl : \textquotedbl unknown\textquotedbl \} \\
] \\
\textasciigrave \textasciigrave \textasciigrave  \\
 \\
Now generate the JSON list based on the given video: \\

\bottomrule[1pt]
\caption{Prompt used for automatic speaker identification.}
\label{tab:prompt_speaker_identification}
\end{longtable}

\section{Reward Design in Phase One}\label{appendix:reward_details}
The reward components are already scaled in their definitions and are summed directly to obtain $r_{\mathrm{mc}}$. We provide the detailed reward definitions used in Phase One training. The reward is computed with a format-gated procedure. If the generated trajectory violates the required output format, it receives only a format penalty, with all other reward components set to zero. Otherwise, it is evaluated using the thinking length, quality, and richness rewards. This hierarchical design treats format correctness as a prerequisite, since malformed outputs cannot be reliably parsed or evaluated. For format-valid outputs, the quality reward serves as the primary learning signal, encouraging memories to be factually grounded, contextually coherent, non-redundant, well-formed, and compliant with the memory budget. The thinking-length reward discourages unnecessarily long intermediate reasoning traces, while the richness reward provides a small auxiliary signal for high-quality candidates, favoring non-redundant, content-rich episodic information. This prevents the policy from exploiting the quality reward by producing short but uninformative memories.

\noindent\textbf{Format reward.}
The model is required to output an intermediate reasoning trace and a final memory in a predefined structure. The reasoning trace must be enclosed within the required thinking tags, and the final memory must follow the specified output schema. We use a binary format reward:
\begin{equation}
r_{\mathrm{fmt}}(\tau)=
\begin{cases}
0, & \text{if the output format is valid},\\
-1.5, & \text{otherwise}.
\end{cases}
\end{equation}

\noindent\textbf{Thinking length reward.}
The thinking length reward is applied only when the output format is valid. It penalizes overlong intermediate reasoning traces, which helps control computational overhead and discourages the policy from allocating excessive tokens to reasoning rather than memory content. Let $\ell_{\mathrm{think}}$ denote the token length of the reasoning trace, and let $L_{\mathrm{think}}$ be the length threshold. We define:
\begin{equation}
\label{eq:r_len}
r_{\mathrm{len}}(\tau)=
\begin{cases}
0, & \ell_{\mathrm{think}} \le L_{\mathrm{think}},\\[4pt]
\lambda_{\mathrm{len}} \dfrac{\ell_{\mathrm{think}}-L_{\mathrm{think}}}{L_{\mathrm{think}}},
& L_{\mathrm{think}} < \ell_{\mathrm{think}} < 2L_{\mathrm{think}},\\[8pt]
\lambda_{\mathrm{len}}, & \ell_{\mathrm{think}} \ge 2L_{\mathrm{think}},
\end{cases}
\end{equation}
where $\lambda_{\mathrm{len}}=-1.0$. In all experiments, we set
$L_{\mathrm{think}}=1200$.

\noindent\textbf{Quality reward.}
The quality reward is the primary learning signal for Phase One. It evaluates whether the generated memory is factually grounded in the video, coherent with the previous memory context, non-redundant, textually well-formed, and compliant with the memory token budget. A candidate is considered quality-valid only if it satisfies all quality criteria. We use a binary quality reward:
\begin{equation}
r_{\mathrm{qual}}(\tau)=
\begin{cases}
0.5, & \text{if the memory satisfies all quality criteria},\\
-0.5, & \text{otherwise}.
\end{cases}
\end{equation}

\noindent\textbf{Richness reward.}
The richness reward is applied only to candidates that are both format-valid and quality-valid. 
It encourages the policy to generate memories with more non-redundant and content-rich information, rather than short but acceptable outputs.

Given a rollout group $\{\tau_i\}_{i=1}^{G}$, where $\tau_i=(q,m_{k,i})$, let $\mathcal{V}$ denote the subset of valid candidates:
\begin{equation}
\mathcal{V}
=
\left\{
i \in \{1,\ldots,G\}
\mid
r_{\mathrm{fmt}}(\tau_i)=0
\text{ and }
r_{\mathrm{qual}}(\tau_i)\ge \gamma_{\mathrm{qual}}
\right\}.
\end{equation}
Here, $r_{\mathrm{fmt}}(\tau_i)=0$ indicates that the output satisfies the required format, and $\gamma_{\mathrm{qual}}$ is the threshold for quality validity. Here $\gamma_{\mathrm{qual}}$ is set as $0$.

Let $G'=|\mathcal{V}|$. 
For each $i\in\mathcal{V}$, a reward model ranks $m_{k,i}$ by richness among the valid candidates, where $\mathrm{rank}(m_{k,i})=1$ indicates the richest memory. 
When $G'>1$, we convert the rank into a normalized score:
\begin{equation}
u_i = 1-\dfrac{\mathrm{rank}(m_{k,i})-1}{G'-1}.
\end{equation}
The richness reward is then defined as
\begin{equation}
\label{eq:r_rich}
r_{\mathrm{rich}}(\tau_i)
=
\begin{cases}
\lambda_{\mathrm{rich}} u_i, & i\in\mathcal{V} \text{ and } G'>1,\\
0, & \text{otherwise},
\end{cases}
\end{equation}
where $\lambda_{\mathrm{rich}}=0.05$.

\section{Phase One Training Details}
\subsection{Reward Model Implementation}\label{reward_implementation}
The format and thinking length rewards are rule-based. We first parse the model output into an intermediate reasoning trace and a final memory field. The format checker verifies whether the reasoning trace is enclosed within the required thinking tags and whether the final memory is a valid JSON object following the required schema in Table~\ref{tab:prompt_generating_episodic_memory}. If this check fails, the remaining reward components are skipped. Otherwise, the thinking length reward is computed deterministically from the token length of the reasoning trace according to Eq.~\eqref{eq:r_len}.

The quality reward is implemented using external evaluators together with rule-based checks. We use Gemini-2.5-Flash to assess faithfulness to the visual content and contextual coherence, and GPT-4o to check textual validity and non-redundancy. Compliance with the memory token budget is enforced deterministically. A candidate is marked as quality-valid only if it passes all these checks. The prompts are given in Tables~\ref{tab:prompt_correctness} and~\ref{tab:prompt_redundancy}.

The richness reward is implemented by prompting GPT-4o to rank the format-valid and quality-valid memories within each rollout group according to the amount of non-redundant, visually grounded, and content-rich episodic information they preserve. The ranking is converted into scalar rewards according to Eq.~\eqref{eq:r_rich}, with the prompt shown in Table~\ref{tab:prompt_usefulness}. All evaluator prompts are fixed across experiments, and all reward queries use deterministic decoding.

\subsection{Training Hyperparameters of GSPO}\label{app:gspo_setting}
We show the training hyperparameters of GSPO for Phase One in Table~\ref{tab:gspo_hyperparameters}. 
\begin{table}[htbp]
\caption{The hyperparameters used in GSPO training.}
\renewcommand{\arraystretch}{1}
\centering
\scalebox{0.95}{
\begin{tabular}{lc}
\toprule[1pt]
\textbf{Parameter Name} & \textbf{Value} \\
\midrule[0.5pt]
Batch Size & 32  \\
Mini Batch Size & 8  \\
GPU with 80GB memory & 32  \\
Number of Samples in a Group \(G\) & 8 \\
Learning Rate & 1e-6 \\
\bottomrule[1pt]
\end{tabular}
}
\label{tab:gspo_hyperparameters}
\end{table}

\section{Phase Two Training Details}
\label{app:phase_two_training}

\subsection{Rollout Cache Construction}
After the Phase One policy $\pi_0$ is fixed, we precompute a rollout cache for Phase Two training. For each streaming context $q=(v_{1:k},m_{1:k-1})$, where the historical memories are generated by $\pi_0$, we sample $N$ candidate memories from $\pi_0$ for the current clip and store
$\langle q,\mathcal{Y}(q)\rangle$, where
$\mathcal{Y}(q)=\{m_{k,j}\}_{j=1}^{N}$.
This cache is constructed once and kept fixed during Phase Two. When adapting to a task environment, we use the recent questions observed in that environment to construct task-relevance preferences over the cached candidates, without resampling candidates or using ground-truth answers.

\subsection{Task-Relevance Reward Model}
We define the Phase Two task reward model as a task-relevance pairwise preference evaluator.
Given a set of recent downstream questions $\mathcal{T}$ from the current environment and two candidate memories $x,y\in\mathcal{Y}(q)$ generated for the same context $q$, the evaluator compares which memory has higher task relevance under the task distribution represented by $T$:
\[
R_{\mathrm{task}}(x,y;\mathcal{T}) \in \{x \succ y,\; y \succ x,\; x \sim y\}.
\]
Here, $x \succ y$ indicates that $x$ is judged more task-relevant than $y$ under $T$, and $x \sim y$ denotes a tie.
To reduce presentation-order bias, each candidate pair is evaluated twice with the order swapped.
We keep a preference only when the two evaluations are consistent and non-tied; otherwise, the comparison is discarded.
Unlike the Phase One rewards, $R_{\mathrm{task}}$ is not used as an absolute scalar reward, but only provides relative preferences between candidate memories.
In implementation, we instantiate the evaluator by prompting GPT-4o to compare the task relevance of two candidate memories for supporting the questions in $\mathcal{T}$, without access to ground-truth answers.
The corresponding prompt is provided in Table~\ref{tab:prompt_relevance}.

\subsection{Preference Data Construction}

For each context $q$, we construct one DPO training pair $(q,m_k^w,m_k^l)$ from the cached candidate memories $\mathcal{Y}(q)$. The selected pair should satisfy two requirements. First, the two memories should have a clear difference in task relevance under the recent task set $\mathcal{T}$. Second, they should avoid substantial differences in basic memory quality, so that the preference mainly reflects task relevance rather than general memory quality differences.

We construct the preference data in two steps. First, we use $R_{\mathrm{task}}$ to compare all candidate-memory pairs and keep only reliable win/loss results, discarding tied or order-inconsistent comparisons. Second, for each context, we aggregate these pairwise preferences into a directed graph over candidate memories, where an edge $y \rightarrow x$ means that $x$ is preferred to $y$ for the current task distribution. Pairwise preferences may be non-transitive, leading to cycles in the raw graph. We remove comparisons involved in cycles and retain the resulting DAG. A longer path $y \leadsto x$ in this DAG indicates a larger task-relevance gap between $x$ and $y$.

We then select the final DPO pair from the DAG. For a valid pair $\langle x,y\rangle$, we require that $x$ is preferred to $y$, that the preferred memory does not have lower basic quality than the dispreferred memory, i.e., $r_{\mathrm{qual}}(x)\ge r_{\mathrm{qual}}(y)$, and that the two memories are not both low-quality. Among all valid pairs, we choose the one with the largest path distance as the final preference pair. If multiple pairs have the same distance, we choose the pair with the smaller memory-length difference. The selected pair is recorded as $(q,m_k^w,m_k^l)$.

In our implementation, we sample $N=8$ candidate memories for each context. After filtering, 29.17\% of sampled contexts yield valid preference pairs, covering roughly 100 videos for Phase Two training.

\subsection{Training Metric Implementation}\label{app:phase_two_metric}
\noindent\textbf{Accuracy.}
Accuracy measures whether Phase Two preserves the fundamental fidelity of memory generation. It is implemented using Gemini-2.5-Flash together with a rule-based length check. Gemini-2.5-Flash judges whether the generated memory is faithful to the visual content and subtitles of the current video clip, while the length check verifies compliance with the predefined memory token budget $L_{\mathrm{mem}}$. A sample is counted as correct only if it passes both checks. The corresponding prompt is given in Table~\ref{tab:prompt_correctness}.

\noindent\textbf{Non-redundancy rate.}
Non-redundancy measures whether Phase Two preserves the ability to generate incremental memory rather than repeating historical content. It is implemented using GPT-4o, which compares the generated memory against the historical memories and determines whether it introduces new information without repeating previously stored content. The evaluator also checks whether the memory remains textually well-formed and consistent with the expected style. A sample is counted as non-redundant if GPT-4o returns a positive judgment. The corresponding prompt is given in Table~\ref{tab:prompt_redundancy}.

\noindent\textbf{Relevance win/tie/loss ratio.}
Relevance measures whether Phase Two improves task-focused memorization relative to the Phase One reference policy. For each validation trajectory and each environment task, we compare the memory generated by the current policy against the memory produced by the reference policy $\pi_0$. We use GPT-4o with the same task-relevance prompt as in the task reward model: the evaluator receives the recent environment questions and the two candidate memories, and judges whether the current policy's memory is more task-relevant, equally task-relevant, or less task-relevant than the reference memory. We report the corresponding Win Ratio, Tie Ratio, and Loss Ratio. The corresponding prompt is given in Table~\ref{tab:prompt_relevance}.

\subsection{Hyperparameters}\label{app:dpo_setting}
Table~\ref{tab:dpo_hyperparameters} lists the hyperparameters used in DPO training. 

\begin{table}[htbp]
\caption{The hyperparameters used in DPO training.}
\renewcommand{\arraystretch}{1}
\centering
\scalebox{0.95}{
\begin{tabular}{lc}
\toprule[1pt]
\textbf{Parameter Name} & \textbf{Value} \\
\midrule[0.5pt]
Batch Size & 64  \\
Mini Batch Size & 64  \\
Number of 80GB GPUs & 32  \\
Learning Rate & 1e-3 \\
\bottomrule[1pt]
\end{tabular}
}
\label{tab:dpo_hyperparameters}
\end{table}

\section{Additional Results}

\begin{table}[htbp]
\caption{Results for each task in VideoMME.}
\renewcommand{\arraystretch}{1}
\centering
\resizebox{0.95\linewidth}{!}{%
\begin{tabular}{lcccccccccccc}
\toprule[1pt]
\multicolumn{1}{c}{\multirow{2.5}{*}{\textbf{Method}}} & \multicolumn{3}{c}{\textbf{Counting Problem}} & \multicolumn{3}{c}{\textbf{Information Synopsis}} & \multicolumn{3}{c}{\textbf{Object Recognition}} & \multicolumn{3}{c}{\textbf{Action Reasoning}} \\
\cmidrule(l){2-4} \cmidrule(l){5-7} \cmidrule(l){8-10} \cmidrule(l){11-13}
& Acc.($\uparrow$) & Cov.  & Prec.($\uparrow$) & Acc.($\uparrow$) & Cov.  & Prec.($\uparrow$) & Acc.($\uparrow$) & Cov.  & Prec.($\uparrow$) & Acc.($\uparrow$) & Cov.  & Prec.($\uparrow$) \\
\midrule[0.5pt]
EgoGPT & 22.3 & 40.9 & 54.4 & 75.6 & 90.0 & 84.0 & 44.7 & 55.0 & 81.2 & 37.1 & 50.5 & 73.6 \\
HippoMM & 25.5 & 55.9 & 45.5 & 75.6 & 91.9 & 82.3 & 53.0 & 65.7 & 80.7 & 38.1 & 53.3 & 71.4 \\
M3-Agent & 30.5 & 65.0 & 46.9 & 90.6 & 97.5 & 92.9 & 65.3 & 77.7 & 84.1 & 61.0 & 74.3 & 82.1 \\
Gemini-1.5-Pro & 26.8 & 49.5 & 54.1 & 88.1 & 96.9 & 91.0 & 54.7 & 61.7 & 88.6 & 71.4 & 84.8 & 84.3 \\
Gemini-2.5-Pro & 34.5 & 60.5 & 57.1 & 91.9 & 98.1 & 93.6 & 65.7 & 74.3 & 88.3 & 73.3 & 83.8 & 87.5 \\
GPT-5.2 & 39.5 & 68.2 & 58.0 & 91.2 & 97.5 & 93.6 & 68.0 & 80.7 & 84.3 & 74.3 & 84.8 & 87.6 \\
Qwen3-VL-30B-A3B & 34.5 & 60.0 & 57.6 & 90.0 & 98.1 & 91.7 & 62.0 & 72.7 & 85.3 & 61.0 & 76.2 & 80.0 \\
\sysname & 42.3 & 64.1 & 66.0 & 93.1 & 98.8 & 94.3 & 69.7 & 78.7 & 88.6 & 69.5 & 82.9 & 83.9 \\
\bottomrule[1pt]

& \multicolumn{3}{c}{\textbf{Object Reasoning}} & \multicolumn{3}{c}{\textbf{Temporal Perception}} & \multicolumn{3}{c}{\textbf{Attribute Perception}} & \multicolumn{3}{c}{\textbf{Temporal Reasoning}} \\
\cmidrule(l){2-4} \cmidrule(l){5-7} \cmidrule(l){8-10} \cmidrule(l){11-13}
& Acc.($\uparrow$) & Cov.  & Prec.($\uparrow$) & Acc.($\uparrow$) & Cov.  & Prec.($\uparrow$) & Acc.($\uparrow$) & Cov.  & Prec.($\uparrow$) & Acc.($\uparrow$) & Cov.  & Prec.($\uparrow$) \\
\midrule[0.5pt]
EgoGPT & 48.6 & 65.0 & 74.8 & 59.2 & 77.6 & 76.3 & 43.6 & 54.9 & 79.4 & 45.3 & 69.8 & 65.0 \\
HippoMM & 52.8 & 71.0 & 74.3 & 51.0 & 73.5 & 69.4 & 50.8 & 62.6 & 81.1 & 46.5 & 79.1 & 58.8 \\
M3-Agent & 75.2 & 87.4 & 86.1 & 75.5 & 87.8 & 86.0 & 60.0 & 68.7 & 87.3 & 58.1 & 87.2 & 66.7 \\
Gemini-1.5-Pro & 66.4 & 78.0 & 85.0 & 61.2 & 75.5 & 81.1 & 35.9 & 39.5 & 90.9 & 68.6 & 84.9 & 80.8 \\
Gemini-2.5-Pro & 72.4 & 85.0 & 85.2 & 77.6 & 87.8 & 88.4 & 46.2 & 53.3 & 86.5 & 69.8 & 86.0 & 81.1 \\
GPT-5.2 & 72.4 & 86.9 & 83.3 & 79.6 & 93.9 & 84.8 & 62.1 & 71.3 & 87.1 & 67.4 & 88.4 & 76.3 \\
Qwen3-VL-30B-A3B & 68.7 & 80.8 & 85.0 & 73.5 & 89.8 & 81.8 & 54.4 & 64.6 & 84.1 & 70.9 & 87.2 & 81.3 \\
\sysname & 74.3 & 86.4 & 85.9 & 81.6 & 93.9 & 87.0 & 65.6 & 71.3 & 92.1 & 72.1 & 84.9 & 84.9 \\
\bottomrule[1pt]

& \multicolumn{3}{c}{\textbf{Action Recognition}} & \multicolumn{3}{c}{\textbf{OCR Problems}} & \multicolumn{3}{c}{\textbf{Spatial Perception}} & \multicolumn{3}{c}{\textbf{Spatial Reasoning}} \\
\cmidrule(l){2-4} \cmidrule(l){5-7} \cmidrule(l){8-10} \cmidrule(l){11-13}
& Acc.($\uparrow$) & Cov.  & Prec.($\uparrow$) & Acc.($\uparrow$) & Cov.  & Prec.($\uparrow$) & Acc.($\uparrow$) & Cov.  & Prec.($\uparrow$) & Acc.($\uparrow$) & Cov.  & Prec.($\uparrow$) \\
\midrule[0.5pt]
EgoGPT & 34.4 & 47.2 & 72.9 & 48.0 & 62.4 & 76.9 & 43.1 & 58.8 & 73.3 & 66.7 & 77.8 & 85.7 \\
HippoMM & 36.8 & 51.6 & 71.3 & 64.8 & 76.8 & 84.4 & 43.1 & 64.7 & 66.7 & 71.1 & 86.7 & 82.1 \\
M3-Agent & 50.4 & 64.8 & 77.8 & 78.4 & 88.8 & 88.3 & 52.9 & 72.5 & 73.0 & 82.2 & 88.9 & 92.5 \\
Gemini-1.5-Pro & 54.0 & 64.0 & 84.4 & 48.0 & 53.6 & 89.6 & 47.1 & 56.9 & 82.8 & 80.0 & 84.4 & 94.7 \\
Gemini-2.5-Pro & 63.6 & 74.8 & 85.0 & 57.6 & 64.0 & 90.0 & 52.9 & 64.7 & 81.8 & 86.7 & 95.6 & 90.7 \\
GPT-5.2 & 63.6 & 75.2 & 84.6 & 72.0 & 81.6 & 88.2 & 60.8 & 68.6 & 88.6 & 95.6 & 100.0 & 95.6 \\
Qwen3-VL-30B-A3B & 54.0 & 68.0 & 79.4 & 69.6 & 76.8 & 90.6 & 54.9 & 62.7 & 87.5 & 86.7 & 93.3 & 92.9 \\
\sysname & 56.8 & 70.0 & 81.1 & 75.2 & 86.4 & 87.0 & 64.7 & 72.5 & 89.2 & 91.1 & 95.6 & 95.3 \\
\bottomrule[1pt]
\end{tabular}
}
\label{tab:videomme}
\end{table}

\begin{table}[htbp]
\caption{Results for each task in EgoLife.}
\renewcommand{\arraystretch}{1}
\centering
\resizebox{0.95\linewidth}{!}{%
\begin{tabular}{lccccccccccccccc}
\toprule[1pt]
& \multicolumn{3}{c}{\textbf{RelationMap}} & \multicolumn{3}{c}{\textbf{EventRecall}} & \multicolumn{3}{c}{\textbf{EntityLog}} & \multicolumn{3}{c}{\textbf{HabitInsight}} & \multicolumn{3}{c}{\textbf{TaskMaster}} \\
\cmidrule(l){2-4} \cmidrule(l){5-7} \cmidrule(l){8-10} \cmidrule(l){11-13} \cmidrule(l){14-16}
& Acc.($\uparrow$) & Cov.  & Prec.($\uparrow$) & Acc.($\uparrow$) & Cov.  & Prec.($\uparrow$) & Acc.($\uparrow$) & Cov.  & Prec.($\uparrow$) & Acc.($\uparrow$) & Cov.  & Prec.($\uparrow$)  & Acc.($\uparrow$) & Cov.  & Prec.($\uparrow$) \\
\midrule[0.5pt]
EgoGPT & 9.6 & 20.8 & 46.2 & 26.2 & 37.3 & 70.2 & 10.4 & 13.6 & 76.5 & 14.8 & 27.9 & 52.9 & 46.0 & 54.0 & 85.3 \\
HippoMM & 16.8 & 34.4 & 48.8 & 36.5 & 50.8 & 71.9 & 29.6 & 37.6 & 78.7 & 27.9 & 41.0 & 68.0 & 49.2 & 60.3 & 81.6 \\
M3-Agent & 2.4 & 12.8 & 18.8 & 34.1 & 46.8 & 72.9 & 19.2 & 24.8 & 77.4 & 19.7 & 26.2 & 75.0 & 42.9 & 50.8 & 84.4 \\
Gemini-1.5-Pro & 23.2 & 40.0 & 58.0 & 46.0 & 59.5 & 77.3 & 28.0 & 36.8 & 76.1 & 52.5 & 59.0 & 88.9 & 68.3 & 81.0 & 84.3 \\
Gemini-2.5-Pro & 36.8 & 53.6 & 68.7 & 50.8 & 65.1 & 78.0 & 33.6 & 43.2 & 77.8 & 44.3 & 52.5 & 84.4 & 63.5 & 76.2 & 83.3 \\
GPT-5.2 & 13.6 & 27.2 & 50.0 & 42.1 & 60.3 & 69.7 & 30.4 & 40.0 & 76.0 & 42.6 & 52.5 & 81.2 & 63.5 & 77.8 & 81.6 \\
Qwen3-VL-30B-A3B & 30.4 & 49.6 & 61.3 & 42.9 & 57.1 & 75.0 & 24.0 & 36.0 & 66.7 & 54.1 & 60.7 & 89.2 & 58.7 & 73.0 & 80.4 \\
\sysname & 37.6 & 54.4 & 69.1 & 50.0 & 61.1 & 81.8 & 28.0 & 36.8 & 76.1 & 60.7 & 63.9 & 94.9 & 71.4 & 82.5 & 86.5 \\

\bottomrule[1pt]
\end{tabular}
}
\label{tab:egolife}
\end{table}

\begin{table}[htbp]
\caption{Results for each task in EgoTempo.}
\renewcommand{\arraystretch}{1}
\centering
\resizebox{0.95\linewidth}{!}{%
\begin{tabular}{lccccccccccccccc}
\toprule[1pt]
& \multicolumn{3}{c}{\textbf{Action-Specific Object}} & \multicolumn{3}{c}{\textbf{Action Sequence}} & \multicolumn{3}{c}{\textbf{Locating Object}} & \multicolumn{3}{c}{\textbf{Spatial Relationship}} & \multicolumn{3}{c}{\textbf{Object Sequence}} \\
\cmidrule(l){2-4} \cmidrule(l){5-7} \cmidrule(l){8-10} \cmidrule(l){11-13} \cmidrule(l){14-16}
& Acc.($\uparrow$) & Cov.  & Prec.($\uparrow$) & Acc.($\uparrow$) & Cov.  & Prec.($\uparrow$) & Acc.($\uparrow$) & Cov.  & Prec.($\uparrow$) & Acc.($\uparrow$) & Cov.  & Prec.($\uparrow$)  & Acc.($\uparrow$) & Cov.  & Prec.($\uparrow$) \\

\midrule[0.5pt]
EgoGPT & 13.2 & 26.0 & 50.8 & 22.9 & 54.2 & 42.3 & 10.4 & 24.0 & 43.3 & 6.0 & 10.0 & 60.0 & 23.2 & 60.0 & 38.7 \\
HippoMM & 10.8 & 22.0 & 49.1 & 16.7 & 43.8 & 38.1 & 16.4 & 22.0 & 74.5 & 18.0 & 20.0 & 90.0 & 17.6 & 52.0 & 33.8 \\
M3-Agent & 14.4 & 30.0 & 48.0 & 17.1 & 52.1 & 32.8 & 9.6 & 22.0 & 43.6 & 10.0 & 12.0 & 83.3 & 24.4 & 64.0 & 38.1 \\
Gemini-1.5-Pro & 27.2 & 38.0 & 71.6 & 37.1 & 60.4 & 61.4 & 12.4 & 14.0 & 88.6 & 0.4 & 2.0 & 20.0 & 32.4 & 66.0 & 49.1 \\
Gemini-2.5-Pro & 30.4 & 44.0 & 69.1 & 47.5 & 75.0 & 63.3 & 19.2 & 24.0 & 80.0 & 6.4 & 10.0 & 64.0 & 36.4 & 68.0 & 53.5 \\
GPT-5.2 & 26.4 & 40.0 & 66.0 & 44.6 & 66.7 & 66.9 & 27.2 & 38.0 & 71.6 & 22.0 & 30.0 & 73.3 & 39.2 & 78.0 & 50.3 \\
Qwen3-VL-30B-A3B & 29.2 & 44.0 & 66.4 & 37.9 & 62.5 & 60.7 & 11.6 & 20.0 & 58.0 & 6.4 & 10.0 & 64.0 & 31.6 & 70.0 & 45.1 \\
\sysname & 30.0 & 42.0 & 71.4 & 53.1 & 74.3 & 71.5 & 20.4 & 30.0 & 68.0 & 8.0 & 10.0 & 80.0 & 37.2 & 68.0 & 54.7 \\

\bottomrule[1pt]
& \multicolumn{3}{c}{\textbf{Object-Specific Action}} & \multicolumn{3}{c}{\textbf{Temporal Event Ordering}} & \multicolumn{3}{c}{\textbf{Future Action Prediction}} & \multicolumn{3}{c}{\textbf{Counting Actions}} & \multicolumn{3}{c}{\textbf{Counting Objects}} \\
\cmidrule(l){2-4} \cmidrule(l){5-7} \cmidrule(l){8-10} \cmidrule(l){11-13} \cmidrule(l){14-16}
& Acc.($\uparrow$) & Cov.  & Prec.($\uparrow$) & Acc.($\uparrow$) & Cov.  & Prec.($\uparrow$) & Acc.($\uparrow$) & Cov.  & Prec.($\uparrow$) & Acc.($\uparrow$) & Cov.  & Prec.($\uparrow$)  & Acc.($\uparrow$) & Cov.  & Prec.($\uparrow$) \\
\midrule[0.5pt]
EgoGPT & 17.2 & 26.0 & 66.2 & 8.2 & 20.4 & 40.0 & 22.4 & 60.0 & 37.3 & 12.2 & 22.4 & 54.5 & 14.8 & 32.0 & 46.2 \\
HippoMM & 14.0 & 26.0 & 53.8 & 7.3 & 8.2 & 90.0 & 29.2 & 62.0 & 47.1 & 8.6 & 18.4 & 46.7 & 18.8 & 34.0 & 55.3 \\
M3-Agent & 12.0 & 20.0 & 60.0 & 15.5 & 36.7 & 42.2 & 28.0 & 66.0 & 42.4 & 8.6 & 18.4 & 46.7 & 20.8 & 42.0 & 49.5 \\
Gemini-1.5-Pro & 21.2 & 30.0 & 70.7 & 19.2 & 42.9 & 44.8 & 25.2 & 50.0 & 50.4 & 15.9 & 30.6 & 52.0 & 6.4 & 10.0 & 64.0 \\
Gemini-2.5-Pro & 27.2 & 38.0 & 71.6 & 29.8 & 53.1 & 56.2 & 26.8 & 56.0 & 47.9 & 15.5 & 30.6 & 50.7 & 19.6 & 26.0 & 75.4 \\
GPT-5.2 & 38.0 & 48.0 & 79.2 & 33.1 & 57.1 & 57.9 & 40.4 & 76.0 & 53.2 & 21.6 & 36.7 & 58.9 & 28.8 & 44.0 & 65.5 \\
Qwen3-VL-30B-A3B & 18.8 & 26.0 & 72.3 & 29.8 & 46.9 & 63.5 & 28.4 & 58.0 & 49.0 & 16.7 & 32.7 & 51.2 & 12.8 & 20.0 & 64.0 \\
\sysname & 24.8 & 34.0 & 72.9 & 29.4 & 51.0 & 57.6 & 36.8 & 66.0 & 55.8 & 22.9 & 32.7 & 70.0 & 21.2 & 38.0 & 55.8 \\
\bottomrule[1pt]
\end{tabular}
}
\label{tab:egotempo}
\end{table}

\subsection{Robustness to the Choice of Answer Generator} \label{app:robustness}

\begin{table}[t]
\caption{VideoMME results under different answer generators. We keep the generated memories fixed and only replace the model used to answer questions from memory.}
\renewcommand{\arraystretch}{0.95}
\centering
\resizebox{0.7\linewidth}{!}{%
\begin{tabular}{llccc}
\toprule[1pt]
Answer Generator & Model & Acc.($\uparrow$) & Cov.  & Prec.($\uparrow$) \\
\midrule
\multirow{2}{*}{GPT-4o} 
& Qwen3-VL-30B-A3B      & 61.6 & 74.7 & 82.5 \\
& \sysname              & 67.9 & 79.3 & 85.6 \\
\midrule
\multirow{2}{*}{Gemini-2.5-Pro} 
& Qwen3-VL-30B-A3B      & 68.9 & 81.7 & 84.4 \\
& \sysname              & 74.5 & 85.8 & 86.9 \\
\bottomrule[1pt]
\end{tabular}%
}
\label{tab:evaluator_results}
\end{table}

GPT-4o is used in two parts of our pipeline: constructing task-relevance preference data for Phase Two and answering questions from memory during downstream evaluation. This may couple the learned preference signal with the evaluation model. To examine this effect, we conduct an additional robustness check on VideoMME while keeping the generated memories fixed and replacing only the model used for memory-based question answering. This setting preserves the Phase-Two memorization policy but tests whether the learned memories remain useful to an independent answer generator. We compare GPT-4o with Gemini-2.5-Pro under the same memory-based QA protocol.

As shown in Table~\ref{tab:evaluator_results}, changing the answer generator affects the absolute scores, but the relative improvement of \sysname over Qwen3-VL-30B-A3B remains consistent. \sysname improves accuracy by 6.3 points with GPT-4o and by 5.6 points with Gemini-2.5-Pro. Coverage and precision also improve under both generators, indicating that the gains are not tied to a specific QA model.

The higher absolute scores under Gemini-2.5-Pro mainly come from higher coverage: given the same memories, Gemini-2.5-Pro more often identifies sufficient evidence and attempts an answer. Its precision also remains comparable, indicating that the higher coverage is not simply accompanied by indiscriminate answering. Since the generated memories are fixed in this comparison, these differences reflect the behavior of the answer generator rather than changes in memory quality. Despite these generator-dependent differences, \sysname consistently improves over the base memorization policy under both answer generators, suggesting that its relative advantage is robust to the choice of answer generator.

\subsection{Case Studies of Richness Reward}\label{app:case_phase_one}

\begin{longtable}{@{}p{\textwidth}@{}}
\caption{
Comparison of memory generations between \texttt{FOMO} and \texttt{w/o richness reward}.
\hl{Yellow highlights} indicate additional details provided by our model. For privacy protection, human faces in the illustrative video frames are masked only for visualization in this paper; the model is trained and evaluated on the original unmasked videos.
}
\label{table:case_study_memory_cases}
\\

\toprule[1pt]
\endfirsthead

\toprule[0.5pt]
\endhead

\midrule[0.5pt]
\multicolumn{1}{r}{\textit{(Continued on next page)}} \\
\endfoot

\bottomrule[1pt]
\endlastfoot

\textbf{Case 1} \\[3pt]

\texttt{Video} (Illustrated as 7 frames) \\[2pt]
\includegraphics[width=\linewidth]{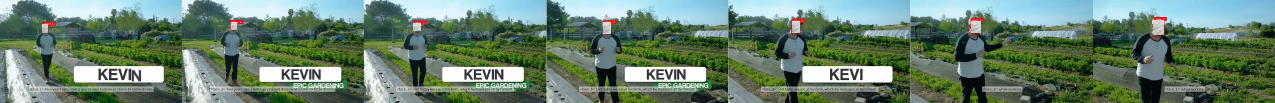}
\\[6pt]

\texttt{Episodic memory (FOMO)} \\[2pt]
\texttt{\textless face\_1\textgreater} walks through a sunlit farm with rows of crops 
\hl{covered in plastic and netting, wearing a white and black long-sleeve shirt, black pants, and a cap}. 
Gesturing with his hands as he speaks, 
\texttt{\textless face\_1\textgreater} says, 
``Five years ago, I met a guy named Andrew at a farm-to-table dinner, 
and today we are at his farm, which he manages all by himself, 
while working 35 hours a week.'' 
\hl{The background includes greenery, a wooden shed, and greenhouses under a clear blue sky}.
\\[6pt]

\texttt{Episodic memory (w/o richness reward)} \\[2pt]
\texttt{\textless face\_1\textgreater} walks through a sunlit farm with rows of crops, 
gesturing while speaking: 
``Five years ago, I met a guy named Andrew at a farm-to-table dinner, 
and today we are at his farm, which he manages all by himself, 
while working 35 hours a week.''

\\[8pt]
\midrule[0.5pt]
\\[-2pt]

\textbf{Case 2} \\[3pt]

\texttt{Video} (Illustrated as 7 frames) \\[2pt]
\includegraphics[width=\linewidth]{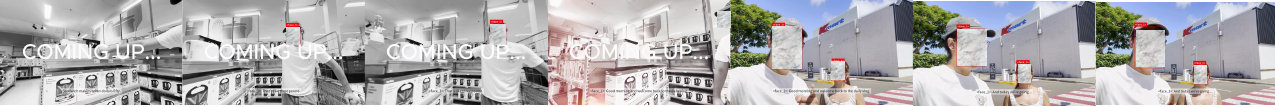}
\\[6pt]

\texttt{Episodic memory (FOMO)} \\[2pt]
\texttt{\textless face\_1\textgreater} stands 
\hl{in a store aisle wearing a cap and white T-shirt, reaching toward shelves with sandwich makers as ``COMING UP...'' text overlays and price commentary play}. 
The scene shifts to outside a Kmart building where 
\texttt{\textless face\_1\textgreater} and 
\texttt{\textless face\_2\textgreater} 
\hl{(in a white sleeveless dress and sunglasses) stand on a crosswalk}; 
\texttt{\textless face\_1\textgreater} gestures while saying, 
``Good morning and welcome back to the daily vlog. 
And today we're going...''
\\[6pt]

\texttt{Episodic memory (w/o richness reward)} \\[2pt]
\texttt{\textless face\_1\textgreater} stands outside a Kmart under a bright blue sky, 
wearing a cap and white T-shirt, while 
\texttt{\textless face\_2\textgreater} stands nearby. 
\texttt{\textless face\_1\textgreater} greets viewers, saying, 
``Good morning and welcome back to the daily vlog. 
And today we're going...''

\\[8pt]
\midrule[0.5pt]
\\[-2pt]

\textbf{Case 3} \\[3pt]

\texttt{Video} (Illustrated as 7 frames) \\[2pt]
\includegraphics[width=\linewidth]{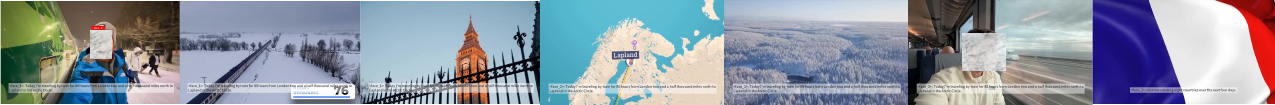}
\\[6pt]

\texttt{Episodic memory (FOMO)} \\[2pt]
\texttt{\textless face\_1\textgreater} 
\hl{stands beside a green train at night, wearing a blue jacket and gray beanie, with snow-covered ground and other people walking in the background}. 
The text states, 
``Today I'm traveling by train for 80 hours from London two and a half thousand miles north to Lapland in the Arctic Circle.'' 
\hl{An aerial view shows a train moving through a snowy landscape with ``76 HOURS REMAINING'' displayed. 
The scene shifts to Big Ben behind a black iron fence, then a map highlighting Lapland with \texttt{\textless face\_1\textgreater}'s image. 
A snowy forest appears, followed by a person in a train looking out the window. 
Finally, the Swedish flag waves as the text reads, ``Join me crossing eight countries over the next four days.''}
\\[6pt]

\texttt{Episodic memory (w/o richness reward)} \\[2pt]
\texttt{\textless face\_1\textgreater} travels by train through snowy landscapes, 
passing landmarks like Big Ben and a map highlighting Lapland, while stating, 
``Today I'm traveling by train for 80 hours from London two and a half thousand miles north to Lapland in the Arctic Circle. 
Join me crossing eight countries over the next four days.''

\end{longtable}

\subsection{Hard Case Analysis}
We conduct an error analysis to investigate the hard cases for our approach. Specifically, we randomly sampled 50 incorrectly answered questions and manually inspected their corresponding generated memories. For each case, we examined whether the memory preserved sufficient evidence to answer the question, whether the preserved evidence was factually accurate, and whether the QA model still failed even when relevant evidence was available.
Across the sampled cases, we identify three major failure modes: (1) missing fine-grained visual evidence, (2) inaccurate or misleading visual descriptions, and (3) insufficient relational and temporal integration. Representative examples are shown in Tables~\ref{tab:case_missing_detail}--\ref{tab:case_missing_control_relation}.

Overall, these findings suggest that future work should incorporate richer visual information into the memory. Key open questions include how to structure and organize visual memory, how to determine which visual details are worth retaining, and how to preserve spatial, relational, and temporal cues from cross-modal information.

\begin{longtable}{@{}p{\textwidth}@{}}
\caption{
Hard case caused by missing fine-grained evidence in the generated memory.
}
\label{tab:case_missing_detail}
\\

\toprule[1pt]
\endfirsthead

\toprule[0.5pt]
\endhead

\midrule[0.5pt]
\multicolumn{1}{r}{\textit{(Continued on next page)}} \\
\endfoot

\bottomrule[1pt]
\endlastfoot

\textbf{Question:} 
What is the number of the first lipstick she used? \\[4pt]

\textbf{Options:} 
A. 959. \quad
B. 656. \quad
C. 858. \quad
D. 666. \\[2pt]

\textbf{Ground-truth Answer:} 
B\\[2pt]

\textbf{Model Prediction:} 
E\\[4pt]

\textbf{Generated Memory.} \\[2pt]

\textbf{00:00:00--00:00:10} \\
The video opens with text reading ``Beauty Secrets'' followed by ``Adriana Lima's Two-Minute Party Makeup.'' 
\facebox{1} appears in a bathroom with tiled walls and a shower area, wearing a black-and-white robe with hair tied back. 
She waves both hands, says ``Hi everyone. My name is Adriana. I love makeup, so this is what I do,'' then adjusts her hair and holds up a BB Cream product labeled ``BB Cream'' while speaking. \\[3pt]

\textbf{00:00:10--00:00:20} \\
\facebox{1} holds up a primer bottle labeled ``Master Prime'' while stating ``Primer,'' then mixes the product on her hand before applying it to her face. 
She explains, ``You see, I mix it up like that. You just go for it. I always like the glow,'' as she spreads the primer across her skin with her fingers, maintaining a focused expression while demonstrating the makeup technique in the bathroom setting. \\[3pt]

\textbf{00:00:20--00:00:30} \\
\facebox{1} continues speaking in the bathroom, stating, ``I don't like when your skin has a matte finish,'' while using hand gestures to emphasize her point. 
She then picks up a white cup, takes a sip, and says, ``English breakfast tea. Okay with honey, that's what I like,'' maintaining a conversational tone as she shares her personal preference for tea while remaining in the tiled bathroom setting. \\[3pt]

\textbf{00:00:30--00:00:40} \\
\facebox{1} holds up a Maybelline ``The Nudes'' eyeshadow palette, saying, ``The nudes, and I'm mixing these two colors.'' 
She uses a brush to blend two shades from the palette, then applies the mixture under her eye with precise hand movements. 
Maintaining a focused expression, she continues explaining the makeup technique while demonstrating the application in the tiled bathroom environment. \\[3pt]

\textbf{00:00:40--00:00:50} \\
\facebox{1} holds a highlighter palette, selecting a shade while stating, ``This one here and then I go.'' 
She applies the highlighter to her cheekbones and forehead with a brush, emphasizing a glowing finish. 
Next, she picks up a mascara wand, says ``Mascara,'' and carefully applies it to her upper lashes, ensuring even coverage. 
Throughout, she maintains a focused demeanor, continuing her makeup tutorial in the tiled bathroom setting. \\[3pt]

\textbf{00:00:50--00:01:00} \\
After applying mascara, \facebox{1} remarks, ``You see my eyes is so weird coming in,'' while adjusting her gaze. 
She then shifts focus to her eyebrows, stating, ``Oh my eyebrows, I'm just gonna leave it naturally and I can just brush with my fingers,'' as she uses her fingers to groom them. 
Next, \facebox{1} picks up a contour palette, declares, ``I'm gonna use the contour,'' and prepares to apply the product, maintaining her instructional tone in the consistent bathroom setting with tiled walls and a shower visible in the background. \\[3pt]

\textbf{00:01:00--00:01:10} \\
\facebox{1} transitions to applying blush, holding up a blush palette and stating, ``a little bit of the blush.'' 
She uses a brush to apply the product to her cheeks, focusing on achieving a natural flush. 
Throughout the process, she maintains her instructional tone, occasionally adjusting her position to ensure even application while remaining in the consistent bathroom setting with tiled walls and a shower visible in the background. \\[3pt]

\textbf{00:01:10--00:01:20} \\
\facebox{1} blends blush on her cheeks with her hand, saying, ``And if it's a little pinkish, I kind of like it,'' while ensuring an even application. 
She then transitions to lipsticks, stating, ``I'm obsessed with lipsticks,'' as she readies the next step in her makeup routine within the tiled bathroom environment. 
\missing{The memory only states that she transitions to lipsticks, but it does not record the visible number or shade label of the first lipstick, which is the key evidence required by the question.} \\[3pt]

\textbf{00:01:20--00:01:30} \\
\facebox{1} holds up a lipstick, saying ``Let's see,'' before applying it to her lips with careful precision. 
She then picks up a lip liner, preparing to define her lip shape while maintaining her focused demeanor. 
Throughout, she remains in the tiled bathroom setting, continuing her makeup tutorial with an instructional tone as she experiments with different lipstick shades to achieve her desired look. \\[3pt]

\textbf{00:01:30--00:01:40} \\
After applying lip liner, \facebox{1} adjusts her hair while stating, ``This can be a cocktail look, guys,'' and emphasizes, ``Look, you just change the lipstick.'' 
She continues refining her lip makeup with deliberate movements, maintaining focus on achieving the desired shade and shape. 
The bathroom setting remains consistent, with tiled walls and a shower visible in the background, as she concludes this segment of her makeup tutorial with a practical demonstration of lipstick customization. \\[3pt]

\textbf{00:01:40--00:01:50} \\
After applying lipstick and exclaiming ``Ta-da,'' \facebox{1} smiles broadly, then says, ``Love you all and see you soon'' while pointing toward the camera. 
The scene concludes with a white screen displaying ``Vogue.com'' as \facebox{1} finishes her makeup tutorial in the consistent bathroom setting with tiled walls and a shower visible in the background. \\[4pt]

\end{longtable}

\begin{longtable}{@{}p{\textwidth}@{}}
\caption{
Hard case caused by incorrect or misleading information in the generated memory.
}
\label{tab:case_wrong_shirt_color}
\\

\toprule[1pt]
\endfirsthead

\toprule[0.5pt]
\endhead

\midrule[0.5pt]
\multicolumn{1}{r}{\textit{(Continued on next page)}} \\
\endfoot

\bottomrule[1pt]
\endlastfoot

\textbf{Question:} 
On which shirt does the man show drawing imaginary lines? \\[2pt]

\textbf{Options:} 
A. The shirt of white. \quad
B. The shirt of black. \quad
C. The shirt of blue. \quad
D. The shirt of yellow. \\[2pt]

\textbf{Ground-truth Answer:} 
B\\[2pt]

\textbf{Model Prediction:} 
C\\[4pt]

\textbf{Generated Memory:} \\[2pt]

\textbf{00:00:00--00:00:10} \\
A white short-sleeve shirt with blue vertical stripes lies flat on a wooden surface. 
Text on the screen reads, ``Today I'm going to show you how to fold a shirt in under two seconds.'' 
A pair of hands enters the frame, swiftly folding the shirt by aligning the sleeves and pressing the fabric. 
The folded shirt is placed back on the surface, neatly arranged with the collar visible. 
New text appears: ``Start by taking a short sleeve shirt'' as the hands adjust the folded garment. \\[3pt]

\textbf{00:00:10--00:00:20} \\
A \wrong{dark blue short-sleeve polo shirt} with a small emblem on the chest is placed flat on the wooden surface. 
The hands adjust the shirt to lie smoothly, ensuring it is properly positioned. 
Text on the screen reads, ``Start by taking a short sleeve shirt and lying it out on its back.'' 
The hands then gesture to indicate imaginary lines: one halfway between the top and bottom of the shirt, and another between the center line and the outside, as the voiceover explains the folding preparation steps. \\[3pt]

\textbf{00:00:20--00:00:30} \\
The hands point to the intersection of the imaginary lines on the \wrong{dark blue polo shirt}, labeling the crosspoint as A, the top as B, and the bottom as C while the voiceover states, ``We'll call the point where the lines cross A, the top point B, and the bottom point C.'' 
The hands then shift to pinch the shirt at point A with the left hand, following the instruction, ``Start by pinching the shirt at point A with your left hand,'' as the preparation for the folding process continues. \\[3pt]

\textbf{00:00:30--00:00:40} \\
The hands continue the folding process by lifting point B with the right hand while maintaining a pinch at point A with the left hand, then crossing the right hand over to grasp point C. 
After securing these points, the hands swiftly unfold the arms and use the wooden surface to fold the shirt backward, aligning the fabric neatly. 
The voiceover instructs, ``Next, quickly unfold your arms, and finally use the table to fold the shirt back,'' as the shirt is transformed into a compact shape through precise hand movements. \\[3pt]

\textbf{00:00:40--00:00:50} \\
The hands place a yellow short-sleeve shirt with a graphic design onto the wooden surface, smoothing it out before folding. 
Following the same technique, the hands pinch point A, lift point B, and cross to grasp point C, then unfold and use the table to fold the shirt neatly. 
Next, a light blue short-sleeve shirt is laid out, adjusted, and folded using the identical method. 
As the final shirt is folded, text appears on screen: ``If you like this video, maybe you'd like to'' while the hands complete the fold, leaving the shirt compact and neatly arranged on the wooden surface. \\[3pt]

\textbf{00:00:50--00:01:00} \\
The video transitions to an end screen featuring a neatly folded blue dress shirt on the left. 
On the right, four video thumbnails are displayed with titles: ``How To Make A Mini Bow And Arrow,'' ``How To Make The Coca-Cola Truck,'' ``How To Make A Light Bulb Vase,'' and ``How To Make A Coca-Cola Can Rose.'' 
Text above the thumbnails reads ``Click To View,'' while below the shirt, the text states, ``If you like this video, maybe you'd like to take a look at some of my others by clicking on the links on the right-hand side.'' 
A yellow ``Subscribe'' button appears at the bottom left, with ``DaveHax.com'' beneath it. 
The message ``Thanks for watching.'' is also visible, concluding the video. \\[3pt]

\textbf{00:01:00--00:01:10} \\
The person, wearing a blue and yellow striped shirt, demonstrates folding a \wrong{dark blue polo shirt} on a wooden surface. 
They mark points A, B, and C on the shirt, pinch at A with the left hand, lift B with the right hand, cross the right hand to grasp C, then unfold and use the table to fold the shirt neatly. 
This process repeats with a yellow graphic short-sleeve shirt and a light blue short-sleeve shirt, each folded using the same technique. 
The video concludes with an end screen featuring a neatly folded blue dress shirt, four video thumbnails titled ``How To Make A Mini Bow And Arrow,'' ``How To Make The Coca-Cola Truck,'' ``How To Make A Light Bulb Vase,'' and ``How To Make A Coca-Cola Can Rose,'' along with text prompting viewers to click for more content, a ``Subscribe'' button, and ``DaveHax.com'' at the bottom. \\[4pt]

\end{longtable}

\begin{longtable}{@{}p{\textwidth}@{}}
\caption{
Hard case where the generated memory captures local visual cues but fails to integrate their relational and temporal structure.
}
\label{tab:case_missing_control_relation}
\\

\toprule[1pt]
\endfirsthead

\toprule[0.5pt]
\endhead

\midrule[0.5pt]
\multicolumn{1}{r}{\textit{(Continued on next page)}} \\
\endfoot

\bottomrule[1pt]
\endlastfoot

\textbf{Question:} 
How does the girl perceive her future with regard to decision-making and control? \\[2pt]

\textbf{Options:} 
A. She doesn't care about her future. \quad
B. She wishes to make decision by herself. \quad
C. She feels she will be controlled by her mom. \quad
D. She has no hope for the future. \\[2pt]

\textbf{Ground-truth Answer:} 
C \\[2pt]

\textbf{Model Prediction:} 
B \\[4pt]

\textbf{Generated Memory:} \\[2pt]

\textbf{00:00:00--00:00:10} \\
\facebox{1} stands against a light blue backdrop, wearing a red shirt and holding a drawing featuring a cartoon figure and hearts. 
The text ``When I try to hide'' appears as \facebox{1} glances toward \facebox{2}. 
\facebox{2}, in a green shirt with short dark hair, is partially visible on the left. 
The text ``there you are'' displays as \facebox{2}'s face becomes fully visible, showing a neutral expression. 
\facebox{1} looks down, shifts to a slight frown, then smiles while still holding the drawing. 
The scene transitions to \facebox{2} holding a purple ``ALGEBRA'' book with a parabola diagram against an orange background, with the text ``you were so much more'' appearing as \facebox{2} holds the book and \facebox{1} is visible from the back on the right.  \\[3pt]

\textbf{00:00:10--00:00:20} \\
The video transitions to a three-panel layout. 
The left panel displays a drawing of a purple-haired pony with hearts and clouds. 
The middle panel shows \facebox{1} in a dark dress with a white collar, holding papers and wearing a somber expression. 
The right panel depicts two figures embracing. 
The scene shifts to two panels: the left shows open books with lined pages, while the right panel features \facebox{1} in a light gray shirt, looking concerned. 
A close-up follows of a drawing with a purple-haired pony, a pencil, and the text ``call me fighter''. 
Finally, \facebox{1} stands against an orange background in a black shirt, pointing with both hands as the text ``call me FIGHTER'' appears in bold letters. \\[3pt]

\textbf{00:00:20--00:00:30} \\
\facebox{1} stands beside a wall where the pony drawing is taped, hands raised in a presenting gesture as the text ``call me LOVER'' appears. 
The scene shifts to \facebox{1} in a black shirt against an orange background, making peace signs with both hands while the text ``call me LOVER'' is displayed in large letters. 
Next, \facebox{1} holds multiple books with a neutral expression, and the text ``a drink OR'' shows up. 
The text then changes to ``I'll take you'' as \facebox{1}'s expression softens into a smile. 
Finally, \facebox{1} appears with light blonde hair and a black shirt, standing before an orange backdrop with red puppet strings; one hand is raised as if controlling the strings, with the text ``I'll take you'' still visible. \\[3pt]

\textbf{00:00:30--00:00:40} \\
The video transitions to a brown background where \facebox{1} stands beside a hand-drawn sign reading ``Thanks to my patrons ssaparova\_ soso\_coaster'' with hearts, and ``Thanks for watching'' below. 
\facebox{1} smiles warmly, pointing at the sign with one hand while the other hand rests by their side, concluding the video with a grateful message.\\

\textbf{Analysis.}
The generated memory captures several local visual cues, but it fails to preserve the key relational and temporal information needed to answer the question. This suggests that the failure is not due to missing low-level perception, but rather insufficient integration of visual cues into task-relevant semantic memory.
\end{longtable}

\section{Prompt Templates for Training and Evaluation}

\begin{longtable}{P}
\toprule[1pt]
\textbf{The prompt for generating episodic memory.}\\
\midrule[0.5pt]

\endfirsthead

\toprule[0.5pt]
\endhead

\midrule[0.5pt]
\multicolumn{1}{r}{\textit{(Continued on next page)}}
\endfoot

\endlastfoot

You are given the following content: \\
- A video with corresponding faces (presented via bounding boxes) and subtitles. \\
- [Description of the preceding part] of the video. This field can be empty. \\
 \\
Using the provided face IDs, write a detailed and cohesive description of the given video. The description should capture the complete set of observable and inferable events in the video. Your output should incorporate the following categories (but is not limited to these): \\
 \\
1. Characters\textquotesingle \ Appearance: Describe the characters\textquotesingle \ appearance, such as their clothing, facial features, or any distinguishing characteristics. \\
2. Characters\textquotesingle \ Actions \& Movements: Describe specific gestures, movements, or interactions performed by the characters. \\
3. Characters\textquotesingle \ Spoken Dialogue: Transcribe or summarize what is spoken by the characters. \\
4. Characters\textquotesingle \ Contextual Behavior: Describe the characters\textquotesingle \ roles in the scene or their interaction with other characters, focusing on their behavior, emotional state, or relationships. \\
 \\
Strict Requirements: \\
- If a character has an associated face ID in the video, refer to them ONLY using that face ID. \\
- If characters DO NOT have associated face IDs in the whole video, it\textquotesingle s ok not to describe them. \\
- A character may have multiple face IDs, and the ID currently displayed on the screen should be used for description. \\
- Ensure the continuity and uniformity of content between adjacent descriptions. \\
- Directly describe the video content, DO NOT start with \textquotesingle The video ...\textquotesingle . \\
- If the video has an incomplete ending plot, the last line is truncated or asr is not a complete sentence, DO NOT describe it. \\
- The final output must be a dictionary, with the key being \textquotedbl description\textquotedbl .\\
 \\
Output format: \\
\textasciigrave\textasciigrave\textasciigrave json \\
\{ \\
\qquad \textquotedbl description\textquotedbl : \textquotedbl <face\_1> is standing outside under a blue sky with clouds. <face\_1> gets out of the car and says: \textquotesingle Hello everyone, welcome to my channel\textquotesingle .\textquotedbl  \\
\} \\
\textasciigrave\textasciigrave\textasciigrave \\
 \\
 
[Description of the preceding part]: \\
\{preceding\_descriptions\} \\
 \\
- Generate subsequent descriptions not covered in [Description of the preceding part], maintain coherence with it, and avoid any repetition of similar information. \\
- If [Description of the preceding part] is empty, describe the video from scratch. \\
- Generate the description briefly in one or two sentences. \\
Please output the description.\\

\bottomrule[1pt]
\caption{The prompt for generating episodic memory.}
\label{tab:prompt_generating_episodic_memory}
\end{longtable}

\begin{longtable}{P}
\toprule[1pt]
\textbf{The prompt for generating episodic memory with task prompts.}\\
\midrule[0.5pt]

\endfirsthead

\toprule[0.5pt]
\endhead

\midrule[0.5pt]
\multicolumn{1}{r}{\textit{(Continued on next page)}}
\endfoot

\endlastfoot

You are given the following content: \\
- A video with corresponding faces, presented via bounding boxes, and subtitles. \\
- [Description of the preceding part] of the video. This field can be empty. \\
\\

Using the provided face IDs, write a detailed and cohesive description of the given video. The description should capture the complete set of observable and inferable events in the video. Your output should incorporate the following categories, but is not limited to these: \\
\\

1. Characters\textquotesingle\ Appearance: Describe the characters\textquotesingle\ appearance, such as their clothing, facial features, or any distinguishing characteristics. \\
2. Characters\textquotesingle\ Actions \& Movements: Describe specific gestures, movements, or interactions performed by the characters. \\
3. Characters\textquotesingle\ Spoken Dialogue: Transcribe or summarize what is spoken by the characters. \\
4. Characters\textquotesingle\ Contextual Behavior: Describe the characters\textquotesingle\ roles in the scene or their interaction with other characters, focusing on their behavior, emotional state, or relationships. \\
\\

Strict Requirements: \\
- If a character has an associated face ID in the video, refer to them ONLY using that face ID. \\
- If characters DO NOT have associated face IDs in the whole video, it\textquotesingle s ok not to describe them. \\
- A character may have multiple face IDs, and the ID currently displayed on the screen should be used for description. \\
- Ensure the continuity and uniformity of content between adjacent descriptions. \\
- Directly describe the video content, DO NOT start with \textquotesingle The video ...\textquotesingle . \\
- If the video has an incomplete ending plot, the last line is truncated, or ASR is not a complete sentence, DO NOT describe it. \\
- The final output must be a dictionary, with the key being \textquotedbl description\textquotedbl . \\
\\

- Your generated memory will be used to answer questions of the same type as those below. Please describe as clearly as possible the information in the video that helps answer such questions. \\
\qquad - \{example\_question\_1\} \\
\qquad - \{example\_question\_2\} \\
\qquad - \{example\_question\_3\} \\
\qquad - \{...\} \\
\\

Output format: \\
\textasciigrave\textasciigrave\textasciigrave json \\
\{ \\
\qquad \textquotedbl description\textquotedbl : \textquotedbl <face\_1> is standing outside under a blue sky with clouds. <face\_1> gets out of the car and says: \textbackslash\textquotedbl Hello everyone, welcome to my channel\textbackslash\textquotedbl .\textquotedbl \\
\} \\
\textasciigrave\textasciigrave\textasciigrave \\
\\

[Description of the preceding part]: \\
\{preceding\_descriptions\} \\
\\

- Generate subsequent descriptions not covered in [Description of the preceding part], maintain coherence with it, and avoid any repetition of similar information. \\
- If [Description of the preceding part] is empty, describe the video from scratch. \\
- Generate the description briefly in one or two sentences. \\
Please output the description. \\

\bottomrule[1pt]
\caption{The prompt for generating episodic memory with task prompts.}
\label{tab:prompt_generating_episodic_memory_with_supplement}
\end{longtable}
\begin{longtable}{P}
\toprule[1pt]
\textbf{The prompt for correctness of episodic memory.}\\
\midrule[0.5pt]

\endfirsthead

\toprule[0.5pt]
\endhead

\midrule[0.5pt]
\multicolumn{1}{r}{\textit{(Continued on next page)}}
\endfoot

\endlastfoot

You are provided with a video, a description of its preceding segment, and a generated candidate [Description] for the remaining portion. \\
Your task is to evaluate: \\
1. Whether the candidate description is factually accurate based only on visual content and subtitles (ignore audio). \\
2. Whether it connects coherently and naturally with the preceding description, without using transition words such as \textquotedbl continue\textquotedbl . \\
For any spoken content, verify it solely against the displayed subtitles and disregard audio information. \\
Assign exactly one label: \\
1: Correct — The description that meets all of the above criteria. \\
0: Incorrect — Any description that fails to meet the above criteria. \\
 \\
Output Requirements: Return the result in the following valid JSON format only. Do not generate anything else. \\
\{ \\
\qquad \textquotedbl correctness\_rationale\textquotedbl : \textquotedbl Short explanation for marking this description as 1 or 0\textquotedbl , \\
\qquad \textquotedbl correctness\textquotedbl : 1 or 0  \\
\} \\
 \\
The description of the preceding segment: \\
\{preceding\_descriptions\} \\
 \\
The [Description] to verify: \\
\{descriptions\} \\

\bottomrule[1pt]
\caption{The prompt for correctness of episodic memory.}
\label{tab:prompt_correctness}
\end{longtable}

\begin{longtable}{P}
\toprule[1pt]
\textbf{The prompt for redundancy of episodic memory.}\\
\midrule[0.5pt]

\endfirsthead

\toprule[0.5pt]
\endhead

\midrule[0.5pt]
\multicolumn{1}{r}{\textit{(Continued on next page)}}
\endfoot

\endlastfoot

You are given the [Context] and a candidate description that are describing new events. \\
 \\
Your task is to evaluate whether the candidate description satisfies the following conditions. \\
 \\
Return label=0 if any condition is satisfied, else 1: \\
(1) The description repeats any atomic fact already present in the [Context]. \\
(2) It includes any mention of bounding boxes, coordinates, or detection boxes (e.g., \textquotedbl bounding box\textquotedbl , \textquotedbl bbox\textquotedbl , \textquotedbl x1,y1,x2,y2\textquotedbl , \textquotedbl rectangle box around\textquotedbl ). \\
(3) It contains meta phrases like: \textquotedbl subtitles said\textquotedbl , \textquotedbl the subtitles say\textquotedbl , \textquotedbl subtitle reads\textquotedbl , \textquotedbl subtitle says\textquotedbl , or \textquotedbl according to the subtitles\textquotedbl . \\
(4) The quoted speech contains transcript-style speaker labels like \textquotedbl <face\_id> says \textquotesingle<face\_id>: Good.\textquotesingle \textquotedbl  inside quoted dialogue. \\
(5) It includes conclusion-based or context-setting statements such as \textquotedbl this video ends with...\textquotedbl  or \textquotedbl based on previous videos\textquotedbl . \\
 \\
Output Requirements: Return the result in the following valid JSON format only. Do not generate anything else. \\
 \\
\{ \\
\qquad \textquotedbl label\_rationale\textquotedbl : \textquotedbl Short explanation for marking this description as 1 or 0\textquotedbl , \\
\qquad \textquotedbl label\textquotedbl : 1 or 0 \\
\} \\
\\

[Context]: \\
\{preceding\_descriptions\} \\
 \\
candidate description to verify: \\
\{descriptions\} \\

\bottomrule[1pt]
\caption{The prompt for redundancy of episodic memory.}
\label{tab:prompt_redundancy}
\end{longtable}

\begin{longtable}{P}
\toprule[1pt]
\textbf{The prompt for richness of episodic memory.}\\
\midrule[0.5pt]

\endfirsthead

\toprule[0.5pt]
\endhead

\midrule[0.5pt]
\multicolumn{1}{r}{\textit{(Continued on next page)}}
\endfoot

\endlastfoot

You are given a list of descriptions summarized from a video, each associated with a unique ID. Please rank these descriptions based on their usefulness, output their ID from high to low. Usefulness should be determined by the amount of non-redundant, unique information contained in each item; items with more unique and less overlapping information should be ranked higher. Besides, descriptions that include dialogue directly in the narrative (e.g., <face\_id> said, \textquotedbl xxx\textquotedbl ) should be ranked higher than descriptions that reference dialogue by referencing subtitles, captions, or other UI elements. The length of the output list must match the input list exactly.\\
\\
Output format:\\

[RANK START]\\

[2, 1, 3, 6, 4, 5]\\

[RANK END]\\
\\
Input Knowledge:\\
\{descriptions\}\\
\\
Output the list of ID:\\

\bottomrule[1pt]
\caption{The prompt for the richness of episodic memory.}
\label{tab:prompt_usefulness}
\end{longtable}

\begin{longtable}{P}
\toprule[1pt]
\textbf{The prompt for task-relevance pairwise comparison of episodic memories.}\\
\midrule[0.5pt]

\endfirsthead

\toprule[0.5pt]
\endhead

\midrule[0.5pt]
\multicolumn{1}{r}{\textit{(Continued on next page)}}\\
\endfoot

\endlastfoot

You are given two [Description] and some example questions.\\
\\
Based on the focus of the example questions, your task is to evaluate which description contains information that would be more useful for answering similar questions.\\
\\
Output the ID of the more useful description. If both descriptions are equally useful (a tie), output -1.\\
\\
- A set of example questions: \{example\_questions\}\\
- Two [Description]:\\
\{blocks\_text\}\\
\\
Return exactly one JSON object:\\
\{\\
\qquad \textquotedbl more\_useful\_rationale\textquotedbl : \textquotedbl Briefly introduce the reasons for making this judgment\textquotedbl ,\\
\qquad \textquotedbl more\_useful\textquotedbl : \textquotedbl ID of the more useful description or -1\textquotedbl \\
\}\\

\bottomrule[1pt]
\caption{Prompt for task-relevance comparison of episodic memories.}
\label{tab:prompt_relevance}
\end{longtable}

\begin{longtable}{P}

\toprule[1pt]
\textbf{Prompt for VideoMME QA test.}\\
\midrule[0.5pt]
\endfirsthead

\toprule[0.5pt]
\endhead

\midrule[0.5pt]
\multicolumn{1}{r}{\textit{(Continued on next page)}} \\
\endfoot

\bottomrule[1pt]
\caption{The QA test prompts for VideoMME, EgoLife, and EgoTempo.}
\label{tab:qa_test_prompts}
\endlastfoot

Based on the following video description, select one option as the answer to the question. 
Give your reasoning for your answer. 
Output the option letter A, B, C or D. 
If you cannot find the answer, output E. \\
 \\
Video Description: \\
\{memory\_text\} \\
 \\
Question: \\
\{question\_with\_options\} \\
 \\
Reasoning: \\
{[Your reasoning here]} \\
 \\
Answer: {[A|B|C|D|E]} \\

\midrule[0.8pt]
\textbf{Prompt for EgoLife QA test.}\\
\midrule[0.5pt]

Based on the following video description, select one option as the answer to the question. 
The question is asked at the CURRENT time, but the relevant evidence is usually located in the TARGET clips. 
Only output the option letter A, B, C, D or E. 
If you cannot find the answer from the description, use E. \\
 \\
Description: \\
\{descriptions\} \\
 \\
Question: \\
\{question\} \\
 \\
Options: \\
\{options\} \\
 \\
Respond ONLY in strict JSON (no markdown, no code fences, no extra text). \\
The JSON schema is: \\
\{ \\
\qquad \textquotedbl cot\textquotedbl : \textquotedbl Reasoning for the selected answer in English or Chinese\textquotedbl , \\
\qquad \textquotedbl answer\textquotedbl : \textquotedbl A|B|C|D|E\textquotedbl \\
\} \\

\midrule[0.8pt]
\textbf{Prompt for EgoTempo QA test.}\\
\midrule[0.5pt]

These are descriptions of a video that I want to upload, please answer the question. 
You need to answer the question in any case and not demand additional context information. 
Note: All actions mentioned refer to the person recording the video. \\
 \\
Video Description: \\
\{descriptions\} \\
 \\
Question: \\
\{question\} \\
 \\
If the provided description is insufficient to answer the question, output \textquotesingle Insufficient Information\textquotesingle. \\
Answer: \\

\end{longtable}

\end{document}